%% file: lamoo.tex
\documentclass[twoside,11pt]{article}

\usepackage{jmlr2e}
\usepackage{times}
\usepackage{epsfig}
\usepackage{graphicx}
\usepackage{amsmath}
\usepackage{amssymb}
\usepackage{subfig}
\usepackage{booktabs}
\usepackage{balance}
\usepackage{enumitem}
\usepackage{wrapfig}

\usepackage{hyperref}       
\usepackage{url}            
\usepackage{booktabs}       
\usepackage{amsfonts}       
\usepackage{nicefrac}       
\usepackage{microtype}      
\usepackage{xcolor}         

\usepackage{multirow}
\usepackage{wrapfig} 
\usepackage{times}
\usepackage{epsfig}
\usepackage{graphicx}
\usepackage{amsmath}
\usepackage{amssymb}
\usepackage[]{subfig}
\usepackage{algorithm}
\usepackage{algpseudocode}
\usepackage{tabularx}
\usepackage{enumitem}
\usepackage{amsmath}
\usepackage{booktabs}
\usepackage{makecell}
\usepackage{mathrsfs}
\usepackage{xcolor}
\usepackage{xspace}
\usepackage{tablefootnote}
\usepackage[flushleft]{threeparttable}
\usepackage[utf8]{inputenc} 
\usepackage[T1]{fontenc}    

\input{math_command.tex}

\def\ours{\emph{LaMOO}\xspace}
\def\oursnas{\emph{LaMOO}\xspace}

\newcommand{\para }[1]{\emph {\textbf {#1}}}

\renewcommand{\1}{{\em (i)}}
\newcommand{\2}{{\em (ii)}}
\newcommand{\3}{{\em (iii)}}

\ShortHeadings{Multi-objective Neural Architecture Search by Learning Search Space Partitions}{Zhao et.al}
\firstpageno{1}

\begin{document}

\title{Multi-Objective Neural Architecture Search \\ by Learning Search Space Partitions}

\author{\name Yiyang Zhao \email yzhao10@wpi.edu \\
       Worcester Polytechnic Institute\\
       \AND
       \name Linnan Wang \email wangnan318@gmail.com \\
       Brown University\\
        \AND
       \name Tian Guo \email tian@wpi.edu \\
       Worcester Polytechnic Institute\\
       }


\maketitle

\begin{abstract}
Deploying deep learning models requires taking into consideration neural network metrics such as model size, inference latency, and \#FLOPs, aside from inference accuracy. This results in deep learning model designers leveraging multi-objective optimization to design effective deep neural networks in multiple criteria. However, applying multi-objective optimizations to neural architecture search (NAS) is nontrivial because NAS tasks usually have a huge search space, along with a non-negligible searching cost. This requires effective multi-objective search algorithms to alleviate the GPU costs. In this work, we implement a novel multi-objectives optimizer based on a recently proposed meta-algorithm called \ours~\cite{lamoo} on NAS tasks. In a nutshell, \ours speedups the search process by learning a model from observed samples to partition the search space and then focusing on promising regions likely to contain a subset of the Pareto frontier. Using \ours, we observe an improvement of more than 200\% sample efficiency compared to Bayesian optimization and evolutionary-based multi-objective optimizers on different NAS datasets. For example, when combined with \ours, qEHVI achieves a 225\% improvement in sample efficiency compared to using qEHVI alone in NasBench201. For real-world tasks, \ours achieves 97.36\% accuracy with only 1.62M \#Params on CIFAR10 in only 600 search samples. On ImageNet, our large model reaches 80.4\% top-1 accuracy with only 522M \#FLOPs. 
\end{abstract}


\section{Introduction}
\label{sec:introduction}

Nowadays, neural architecture search (NAS) has become instrumental in developing deep learning (DL) models that significantly surpass the performance of hand-crafted models designed by experts~\cite{nasnet, nas-fcos, nas-fpn, alphax, lanas, rea, DARTS, pcdarts, ofa}. Fundamentally, NAS aims to identify the best-performing architectures within a given search space using optimization algorithms such as reinforcement learning~\cite{nasnet, enas, mnasnet}, evolutionary algorithm~\cite{rea, lamonade, nsganetv2, chamnet}, or Bayesian optimization~\cite{lanas, pnas, dppnet}. In real-world deployments, metrics aside from inference accuracy are also valuable for determining a DL model's quality. For example, in face ID recognition and self-driving systems, model designers may pay more attention to the inference latency of the model. In resource-constrained devices, such as NVIDIA drive ORIN for self-driving cars~\cite{orin}, the designer maximizes the model accuracy while minimizing the model size/computational complexity. 

As such, NAS tasks can be formulated as multi-objective optimization problems to automatically design DL models that meet all specified requirements. In this work, we explore the application of a new and effective multi-objective optimizer \ours~\cite{lamoo} on NAS to design superior deep learning models that consider multiple, potentially conflicting, metrics. Briefly, \ours is a generic learning-based approach that effectively partitions the search space for multi-objective optimizations. 
Key details of \ours are presented in \S\ref{sec:lamoo}.
A key question that this work seeks to answer is \emph{how effectively \ours will perform on various multi-objective NAS tasks}.

Mathematically, in multi-objective optimization (MOO) we optimize $M$ objectives $\vf(\vx) = [f_1(\vx), f_2(\vx), \ldots, f_M(\vx)]\in \rr^M$:
\begin{eqnarray}
\min\;& f_1(\vx), f_2(\vx), ..., f_M(\vx)  \label{eq:problem-setting}  \label{prob-formulation} \\
\mathrm{s.t.}\;& \vx \in \Omega \nonumber, 
\end{eqnarray}

where $f_i(\vx)$ denotes the function of objective $i$.

While we could set arbitrary weights for each objective to turn it into a single-objective optimization (SOO) problem, modern MOO methods aim to find the problem's entire \emph{Pareto frontier}, the set of solutions that are not \emph{dominated} by any other feasible solutions. Here we define \emph{dominance} $\vy \prec_\vf \vx$ as $f_i(\vx) \leq f_i(\vy)$ for all functions $f_i$, and exists at least one $i$ s.t. $f_i(\vx) < f_i(\vy)$, $1\le i \le M$. That is, if the condition holds, a solution $\vx$ is always better than solution $\vy$, regardless of how the $M$ objectives are weighted. 
In \S\ref{sec:moo_nas_motivation}, we will show preliminary results where single-objective optimizations fail to produce higher-quality neural architectures compared to multi-objective optimizations.

Multi-objective NAS introduces new challenges to the NAS problem. Multiple objectives bring a more complicated value space due to increasing dimensions. Finding a Pareto set of neural architectures is also more difficult than a single optimal solution. 
To tackle the complexity of the multi-objective NAS problem, we extend \ours to learn promising regions for NAS algorithms as will be described in \S\ref{subsec:integration_diff_nas}.

Specifically, \ours learns to partition the search space into \emph{promising} and \emph{non-promising} regions. Each partitioned region corresponds to a node within a search tree, with the leaf nodes serving as candidates for the search process. Subsequently, \oursnas utilizes two variations of Monte-Carlo Tree Search (MCTS) (as will be described in \S\ref{subsec:search_promising_region_selection}) to select the most promising region based on the Upper Confidence Bound (UCB) values, facilitating new architecture sampling. 
\oursnas requires several initial evaluated architectures, collected via random sampling, to bootstrap the learning and search processes. \oursnas can be integrated with various NAS methods, including one-shot NAS, few-shot NAS, and predictor-based NAS. Details of this integration of \oursnas with NAS are available in \S\ref{subsec:integration_diff_nas}.

Our approach, \oursnas, exhibits superior performance over existing methodologies across multiple NAS benchmarks, which include NAS datasets and real-world deep learning tasks. On the NasBench201 dataset, \oursnas boosts the sample efficiency of qEHVI and CMA-ES by 225\% and 500\%, respectively. Likewise, on the NasBench301 dataset, \ours enhances the sample efficiency of the original qEHVI and CMA-ES by over 200\%. On the HW-NASBench dataset with four different search objectives, \oursnas combined with CMA-ES achieves a search performance increase of over 250\% compared to other baselines. In the context of open-domain NAS tasks, \oursnas also stands out. On the CIFAR-10 image classification task, \oursnas requires 1.5X fewer samples and identifies architectures that not only achieve similar accuracy to state-of-the-art (SoTA) models but also have fewer parameters, at 1.62M \#parameters. On ImageNet,  \oursnas found SoTA models with a top-1 accuracy of 80.4\% at 522 MB \#FLOPs, a top-1 accuracy of 78.0\% at 248 MB \#FLOPs, and a top-1 accuracy of 79.2\% with only 0.57 ms TensorRT latency with FP16 on Nvidia GV100. For the MSCOCO object detection task, \oursnas achieves better performance at 37.6 mAP with fewer \#FLOPs at 109.5G, compared to performance-oriented network ResNet-50~\cite{resnet} as the backbone.

In summary, we make the following main contributions. 
\begin{itemize}
  \item We have extended our previous work, \ours~\cite{lamoo}, to the realm of NAS problems. This study is the first to apply learning space partition in multi-objective NAS tasks. We show that \ours stands as a robust meta-optimizer capable of enhancing multi-objective NAS algorithms.
  \item We introduce a new search strategy called leaf selection to improve the efficiency of selecting promising regions and show that the new search strategy can improve the search efficiency for NAS problems.
  \item We implement \ours on different NAS datasets, including Nasbench201~\cite{nasbench201}, Nasbench301~\cite{nasbench301}, and HW-NASBench~\cite{hwnasbench}. We show that using \ours can improve both Bayesian optimization and evolutionary algorithms by over 200\% to 500\% sample efficiency. 
  \item \ours leads to state-of-the-art architecture performance across most real-world NAS tasks. For instance, on CIFAR-10, \ours achieves a Top-1 accuracy of 97.36\% at 1.62M Parameters. On ImageNet, \ours achieves a Top-1 accuracy of 80.4\% with only 522MB \#FLOPs.

\end{itemize}

\section{Motivation}

\subsection{Learning Space Partitions}
\label{sec:motivation:partition}

\begin{figure}[t]
  \centering
  \includegraphics[width=0.65\columnwidth]{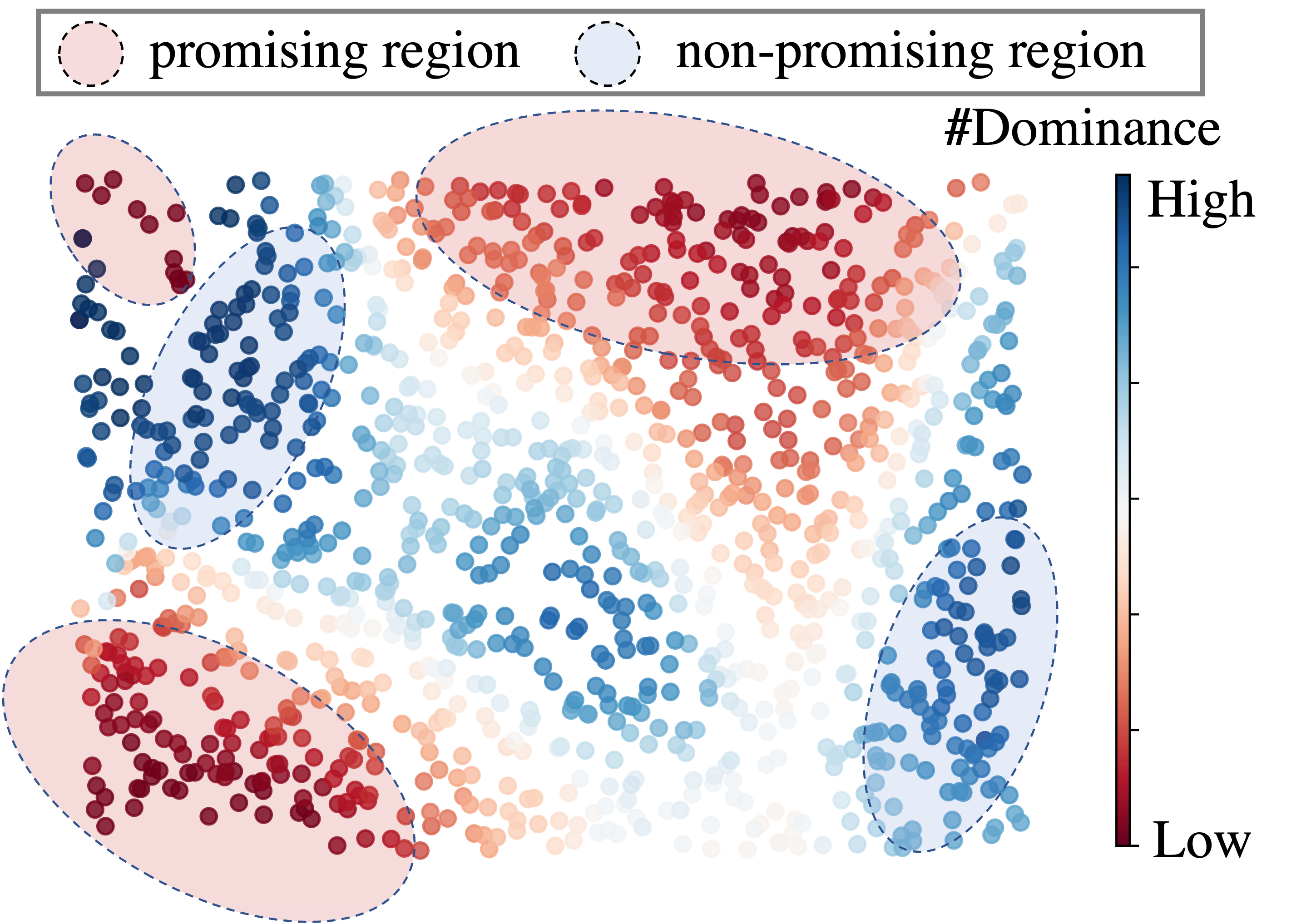}
  \caption{The visualization of search space for the Branin-Currin. Red samples have low dominance numbers, while blue ones have high dominance numbers.
  The smaller the dominance number, the better the quality of the sample.
  }
  \label{fig:motivation}
\end{figure}

In this section, we present a motivating example with the Branin and Currin function~\citep{bc_func}, demonstrating the key benefit of space partitions for multi-objective problems. The Branin-Currin is a 2-dimensional problem with two objectives. 
As described previously, in multi-objective problems, people usually utilize the dominance number to measure the \emph{goodness} of the data point in existing samples~\cite{nsga-ii, nsgaiii}. The dominance number $o(\vx)$ of sample $\vx$ is defined as the number of samples that dominate $\vx$ in search space $\Omega$: 
\begin{equation}
o(\vx) := \sum_{\vx_i \in \Omega}\mathbb{I}[\vx_i \prec_\vf \vx,\ \vx \neq \vx_i],
\label{eq:dominance}
\end{equation}
where $\mathbb{I}[\cdot]$ is the indicator function.
This function indicates that with the decreasing of the $o(\vx)$, $\vx$ would be approaching the Pareto frontier; $o(\vx)$ = $0$ when sample $\vx$ locates in the Pareto frontier. Figure~\ref{fig:motivation} visualize the dominance number of 2000 samples in the search space of the Branin-Currin. Most of the good samples (i.e., small $o(\vx)$) of the search space cluster together in small regions (i.e., shaded by red).

This observation implies that the identification of promising regions and the subsequent concentration of optimization algorithms within these regions can significantly enhance search efficiency. Approaches based on learning space partition~\cite{lanas, lamcts, lamoo} are capable of capitalizing on these benefits. Motivated by this example, and in light of our prior work showcasing \ours as an effective multi-objective optimizer~\cite{lamoo}, our objective in this work is to examine how beneficial search space partition can be for NAS tasks. We approach neural architecture search as a multi-objective problem and apply \ours, our learning partition technique, which will be detailed in \S\ref{sec:lamoo}, to the NAS problem.

\subsection{Multi-Objective Neural Architecture Search}
\label{sec:moo_nas_motivation}

\begin{figure}[t]

\centering 
\subfloat[][constrain: $\#FLOPs$ $<$ 10M]{\includegraphics[width=.48\textwidth]{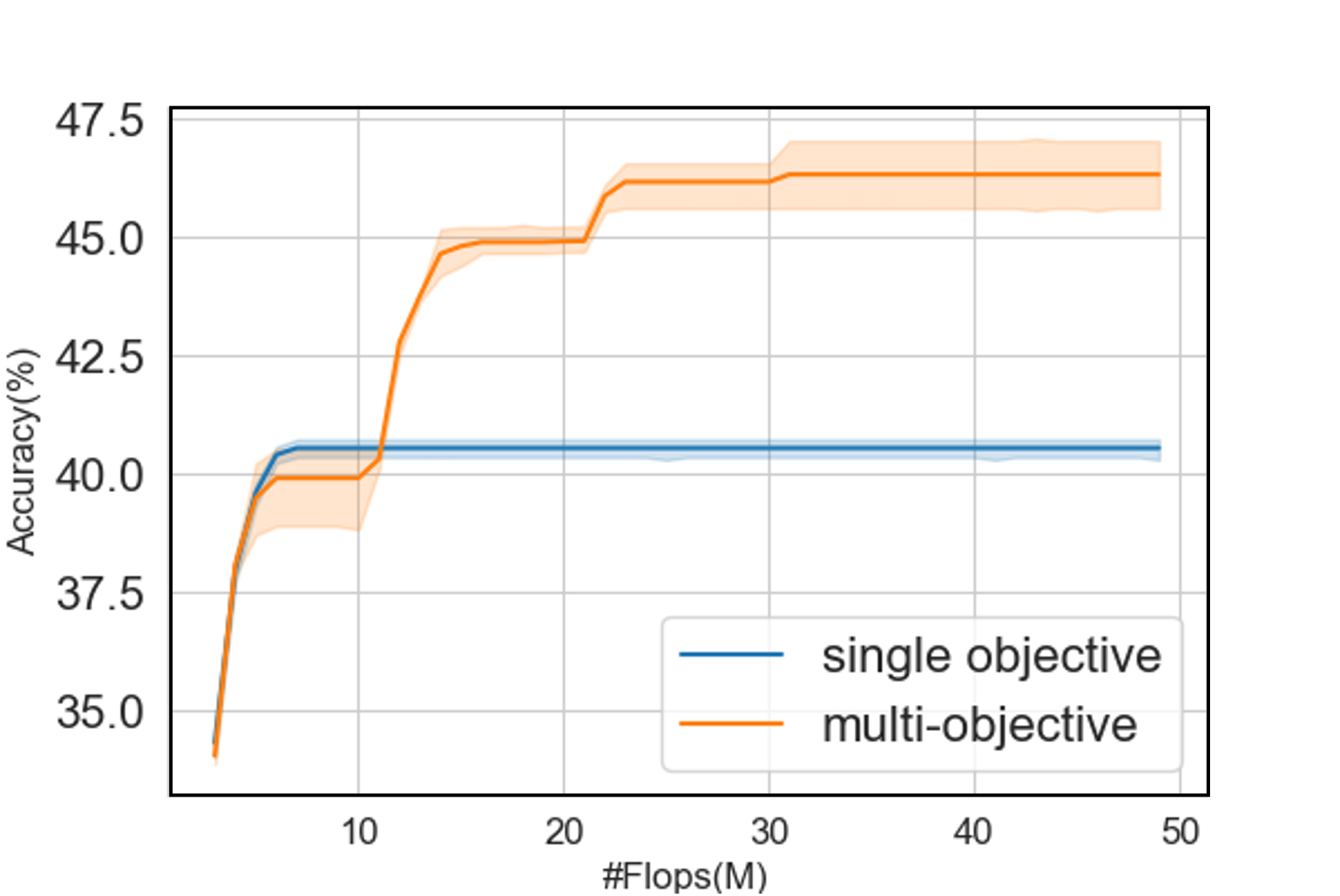}\label{fig:two_a}} \quad
\subfloat[][constrain: $\#FLOPs$ $<$ 30M]{\includegraphics[width=.48\textwidth]{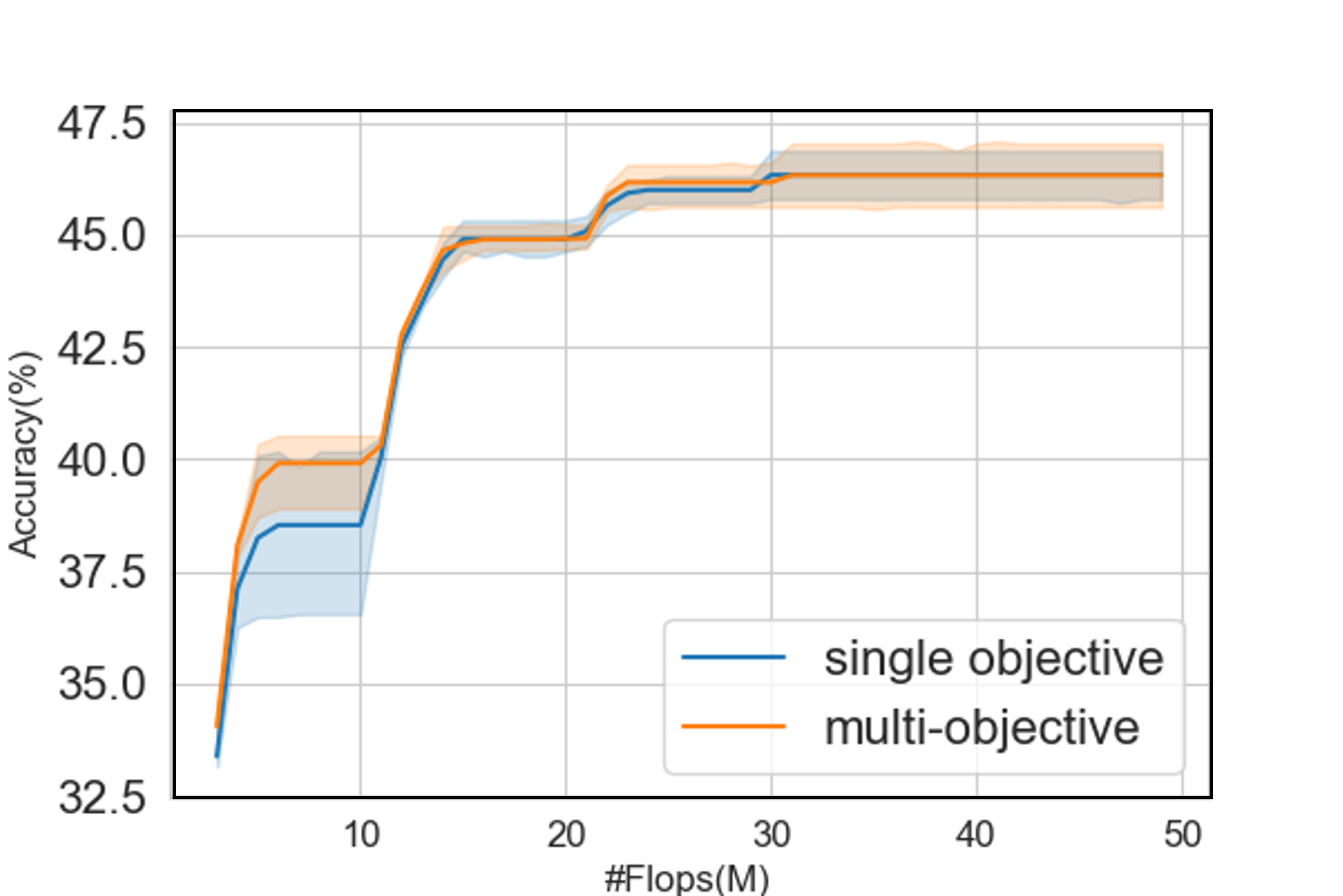}\label{fig:two_b}}  \quad \\
\subfloat[][scalarized objective $\frac{accuracy}{\#FLOPs}$]{\includegraphics[width=.48\textwidth]{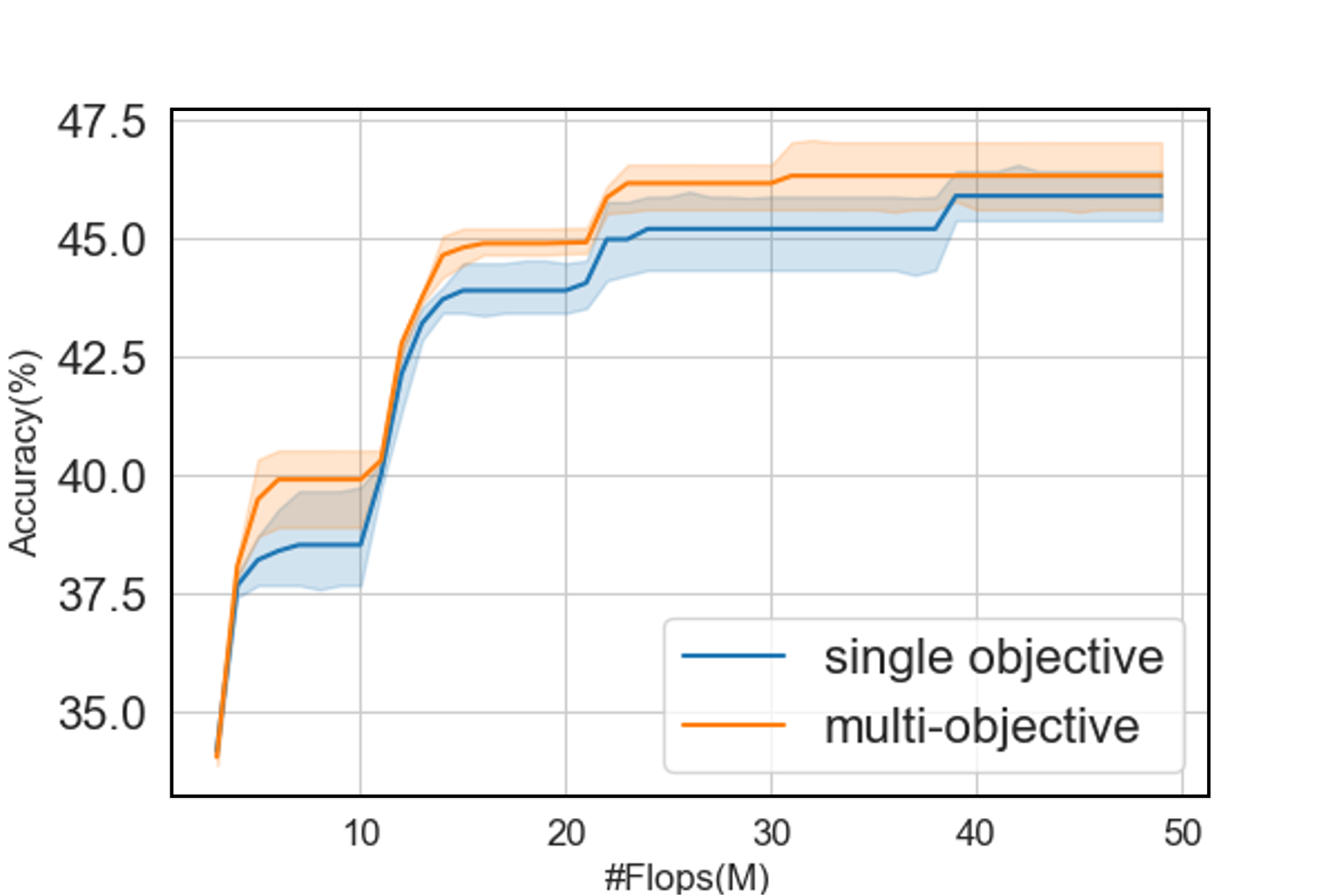}\label{fig:two_c}}  \quad
\subfloat[][contour lines with scalarized objective $\frac{accuracy}{\#FLOPs}$]{\includegraphics[width=.48\textwidth]{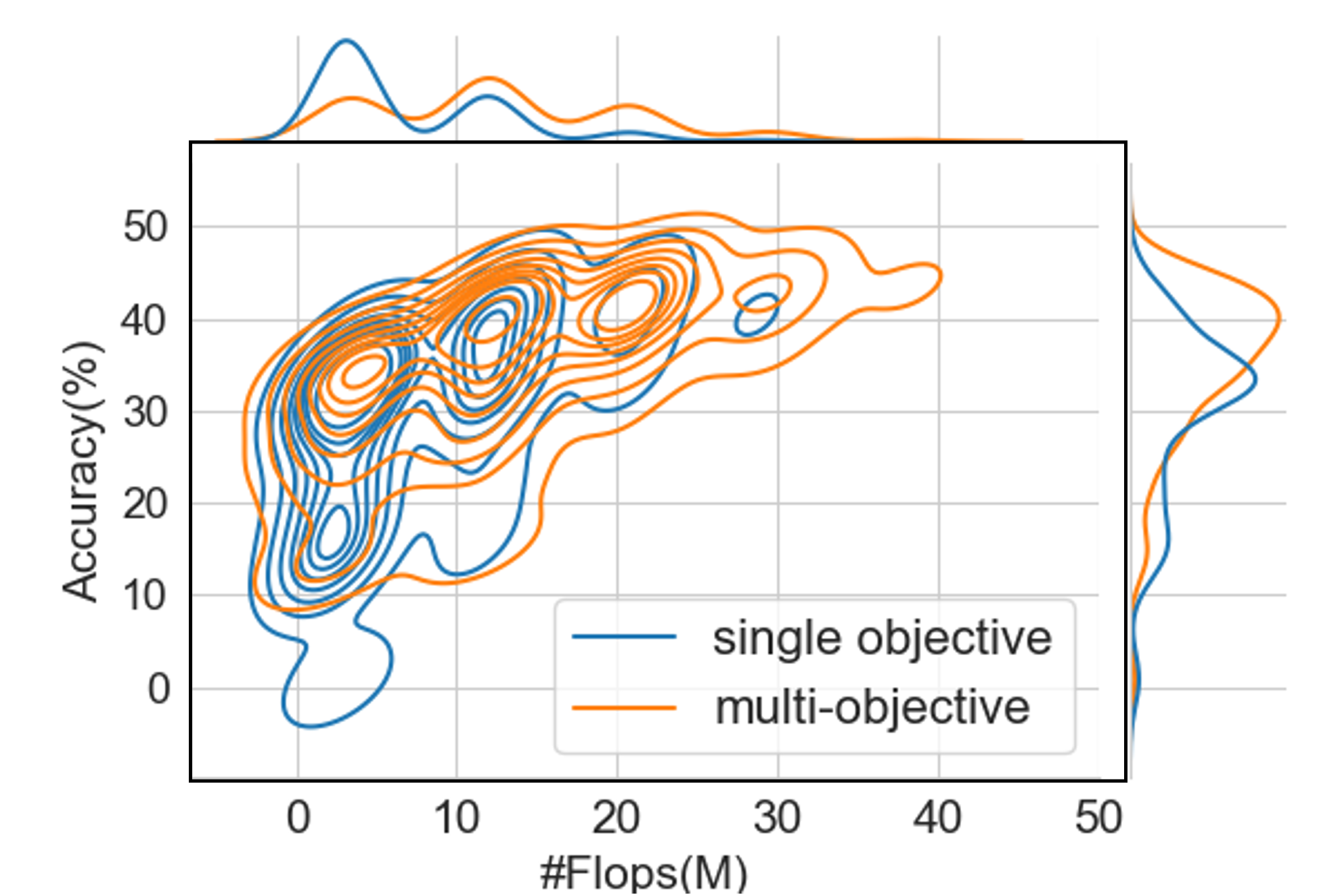}\label{fig:two_d}}  \quad
\caption{500 samples on Nasbench201~\cite{nasbench201} searched by the covariance matrix adaptation evolution strategy (CMA-ES)~\cite{cma, cma-es} with single and multi-objective versions. (a), (b) and (c) plot accuracy vs. \#FLOPs of samples in the Pareto frontier. (d) is the contour lines with a sclarized objective.}
\label{fig:two}
\end{figure}

Efficient deep neural networks concern not only traditional metrics such as accuracy, but also practical efficiency metrics, including inference latency, \#FLOPs, and throughput~\cite{wu2019efficient}. That enables efficient deep neural networks to function in limited compute capacity, limited model complexity, and limited data~\cite{wu2019efficient}. While the design of efficient deep neural networks has drawn increased attention, most NAS works~\cite{fbnet,fbnetv2,proxylessnas,ofa} for efficient deep neural architectures are accuracy-oriented with constraints or by scalarizing different metrics (e.g., $\frac{accuracy}{\#FLOPs}$).

The main limitation of using constraints is that the best samples are only drawn from the constrained regions instead of the global Pareto-frontier. 
For example, if the \#FLOPs is set to be too small, i.e.,  Figure~\ref{fig:two_a}, the single-objective optimization-based search is often limited to architectures with low accuracy. If the \#FLOPs is set to be too large, i.e., Figure.~\ref{fig:two_b}, the single-objective optimization-based search concentrates on architectures of high \#FLOPs. 
This is indicated by the single-objective optimization having significantly worse performance when the \#FLOPS is less than 10M. Moreover, using the scalarized metric may deteriorate the quality of the sampled architectures. As shown in Figure.~\ref{fig:two_c}, multi-objective optimization-based search remains more effective than the single-objective counterpart because MOO better trades off any criterion in search, and Figure.~\ref{fig:two_d} further points out multi-objective search cover more Pareto-set than singe-objective search. 
Our observations suggest that multi-objective NAS is more promising in finding efficient neural architectures with improved search efficiency.

\section{Related Work}
\label{sec:related}
\subsection{Efficient Neural Networks}

Designing neural architectures to achieve the best trade-offs between performance and efficiency has emerged as a popular and important area in the deep learning community in recent years~\cite{mobilenetv1, mobilenetv2, mobilenetv3, ofa, fbnet, fbnetv2, proxylessnas, efficientnet}. In particular, recent innovations focus on designing cost-efficient operations and modules. For example, MobileNetV1~\cite{mobilenetv1} and ShuffleNet~\cite{shufflenet} proposed the depthwise convolution and grouped convolution to reduce the parameters and computations in the traditional convolution operation. MobileNetV2~\cite{mobilenetv2} introduced a cost-friendly inverted residual block (IRB) consisting of an inverted residual and a bottleneck. More recently, the mobile-oriented MobileNetV3~\cite{mobilenetv3} further improves the model performance by using a new h-swish activation and a squeeze and excite module~\cite{semodule} in the IRB. Due to its good performance, IRB has been widely used in state-of-the-art architectures~\cite{mobilenetv2, mobilenetv3, ofa, fbnetv3} as the basic building block. Specifically, IRBs with different parameters and activation types are used to form multiple groups, serving as the key structure of the resulting models. 
In this work, we evaluate the efficiency of \ours in the EfficientNet search space that covers most aforementioned structures and operations. We also consider the connection pattern inside of IRB modules, which is mostly ignored by previous works. We finally compare the resulting efficient neural architectures to the state-of-the-art models.

\subsection{Monte Carlo Tree Search in Neural Architecture Search}

The Monte Carlo Tree Search (MCTS) algorithm is widely used in different areas, such as gaming, robotics planning, optimization, and NAS~\cite{mcts1, mcts2, mcts3, mcts4, lanas, alphax}. AlphaX~\cite{alphax} is the representative of the MCTS-based NAS algorithm. AlphaX directly leverages MCTS to search neural architectures. Each node of MCTS denotes a neural architecture, and the reward of a node is calculated by the architecture's actual performance or a value function predictor. However, MCTS-based NAS agents like AlphaX are unable to deal with multi-objective NAS requirements directly. \ours is a meta-algorithm that leverages MCTS to search the most promising regions for further sampling~\cite{lamoo}, which we apply to optimize for multi-objective NAS problems. We choose to integrate MCTS into \ours because of its effectiveness in balancing exploration and exploitation during the search process~\cite{alphax, mcts1, mcts2}. This integration allows \ours to explore potentially overlooked areas that might contain superior samples, even within regions initially classified as non-promising by \ours. The details of the implementation of MCTS in \ours can be found in \S\ref{sec:monte_carlo_tree_search}. 

Wang et al.~\cite{lanas} was the first to leverage MCTS and partition learning method for \emph{single-objective} NAS problems. On top of the difference between SOO and MOO, the mechanism of the partitioning of the search space between \ours and \cite{lanas} is different. Wang et al.~\cite{lanas} simply uses the median from the single objective of collected samples and learns a linear classifier to separate regions, while \ours leverages dominance rank and an SVM classifier to separate good regions from bad regions.

\subsection{Quality Indicators of Multi-Objective Optimization}

\begin{table}[t]
\centering
\caption{
Comparison of different quality indicators. 
}
\resizebox{0.75\textwidth}{!}{
\begin{tabular}{@{}ccccc@{}}
\toprule
\textbf{Quality Indicator}        & \textbf{Convergence}         & \textbf{Uniformity}          & \textbf{Spread}              & \textbf{No reference set required} \\ \midrule
\multicolumn{1}{|c|}{HyperVolume} & \multicolumn{1}{c|}{$\surd$} & \multicolumn{1}{c|}{$\surd$} & \multicolumn{1}{c|}{$\surd$} & \multicolumn{1}{c|}{$\surd$}       \\
\multicolumn{1}{|c|}{GD}          & \multicolumn{1}{c|}{$\surd$} & \multicolumn{1}{c|}{}        & \multicolumn{1}{c|}{}        & \multicolumn{1}{c|}{}              \\
\multicolumn{1}{|c|}{IGD}         & \multicolumn{1}{c|}{$\surd$} & \multicolumn{1}{c|}{$\surd$} & \multicolumn{1}{c|}{$\surd$} & \multicolumn{1}{c|}{}              \\
\multicolumn{1}{|c|}{MS}          & \multicolumn{1}{c|}{}        & \multicolumn{1}{c|}{}        & \multicolumn{1}{c|}{$\surd$} & \multicolumn{1}{c|}{}              \\
\multicolumn{1}{|c|}{S}           & \multicolumn{1}{c|}{}        & \multicolumn{1}{c|}{$\surd$} & \multicolumn{1}{c|}{}        & \multicolumn{1}{c|}{}              \\
\multicolumn{1}{|c|}{ONVGR}       & \multicolumn{1}{c|}{$\surd$} & \multicolumn{1}{c|}{}        & \multicolumn{1}{c|}{}        & \multicolumn{1}{c|}{}              \\ \bottomrule
\end{tabular}
}
\label{tab:scalarizing_review}
\end{table}

There are several quality indicators~\citep{gd, igd, maxspread, spacing, error_ratio} for evaluating sample quality, which can be used to scalarize the MOO problem to the SOO problem.  The performance of a quality indicator can be evaluated by three metrics~\citep{QI_res1, QI_res2}, including \emph{convergence} (closeness to the Pareto frontier), \emph{uniformity} (the extent of the samples satisfying the uniform distribution), and \emph{spread} (the extent of the obtained samples approximate Pareto frontier). Generational Distance (GD)~\citep{gd} measures the distance between the Pareto frontier of approximation samples and the true Pareto frontier, which requires prior knowledge of the true Pareto frontier, and only considers convergence. Inverted Generational Distance (IGD)~\citep{igd} is an improved version of GD. IGD calculates the distance between the points on the true Pareto frontier to the closest point on the Pareto frontier of current samples. IGD satisfies all three evaluation metrics of QI but requires a true Pareto frontier which is hard to get in real-world problems. Maximum Spread (MS)~\citep{maxspread} computes the distance between the farthest two points of samples to evaluate the spread. Spacing (S)~\citep{spacing} measures how close the distribution of the Pareto frontier of samples is to uniform distribution. Overall Non-dominated Vector Generation and Ratio (ONVGR) is the ratio of the number of samples in the true Pareto frontier. 
In this work, we choose HyperVolume (HV)~\cite{hv} to evaluate the optimization performance of different algorithms because it can simultaneously satisfy the evaluation of convergence, uniformity, and spread without the knowledge of the true Pareto frontier. In addition, HV plays an important role in \ours as we leverage it to identify the goodness of partitioned spaces, from which \ours picks the best one for sampling. More details can be found in \S\ref{sec:learning_phase}. Table~\ref{tab:scalarizing_review} compares the characteristics of each quality indicator.

\subsection{Search Space Optimization for Neural Architecture Search}

In addition to numerous NAS studies that aim to identify the most promising architectures within a given search space, there are a number of works that focus on search space design. These studies~\cite{chen2023towards, xia2022progressive, evolving_space, space_design_1, mcunet} seek to uncover design principles that increase the likelihood of containing more promising architectures. A notable example of this approach is found in \cite{space_design_1}, where the authors designed a straightforward search space termed \emph{AnyNet} by analyzing 500 sampled architectures to identify common traits of successful designs within these samples. However, a critical limitation of this method is that the patterns it identifies are specific to certain datasets/problems/tasks; these patterns may shift when applied to different types of problems, thus limiting its generality and applicability. In contrast, our \oursnas adopts a data-driven approach, systematically narrowing the entire search space into a more promising sub-region based on information collected from previously observed samples. As such, \ours can be applied to any datasets/problems/tasks. We will show that the searched models by \oursnas outperform the ones by  \cite{space_design_1} in Table.~\ref{tab:lamoo_imagenet}. 

MCUNet~\cite{mcunet} offers an alternative method for optimizing the search space in NAS, tailored specifically for neural network architectures that must operate within the constraints of certain devices, such as microcontrollers with limited memory or specific latency requirements. Specifically, MCUNet predefines a variety of search spaces based on different input resolutions and width multipliers. The underlying 
assumption is that models with greater computational requirements have a larger capacity and are, therefore, more likely to achieve higher accuracy. Based on this assumption, MCUNet randomly samples 1000 architectures from each of the predefined search spaces, subsequently selecting those that meet the specific requirements of the target devices (such as memory capacity and latency). By calculating the average \#FLOPs for the architectures that fulfill these criteria within each search space, they identify search spaces with higher average \#FLOPs that have the potential to yield promising architectures. This method, however, has two limitations. 
First, it may not be applicable if the architectural design is constrained by requirements related to \#FLOPs. Second, it requires additional domain knowledge and human effort to design the candidate search spaces. 
Instead of focusing solely on \#FLOPs, our \oursnas can work with multiple metrics (e.g., \#Params, \#FLOPs, latency, accuracy). It utilizes data from previously evaluated architectures to partition the search space and identify promising regions that are more likely to contain architectures meeting the design requirements. \oursnas is a parallel approach to MCUNet~\cite{mcunet}, \oursnas can complement it by further refining the search space based on historical architecture samples. When MCUNet is used to select an appropriate initial search space for a given target device, \oursnas can then refine this space to focus on promising sub-regions.

\section{Multi-Objective Optimization by Learning Space Partition}
\label{sec:lamoo}

\begin{table*}[t]
\caption{Notation definitions through the paper.}
\resizebox{\textwidth}{!}{%
\begin{tabular}{@{}ll|ll|ll@{}}
\toprule
$\Omega$                   & the whole architecture space                 & $a$    & an architecture in the architecture space             & $o(a)$                            & dominance number of architecture $a$      \\
$\Omega_{j}$    & the partition of $\Omega$ represented by the tree node $j$                           & $n_{i}$          & number of samples in node $i$             &  $D_{t} \cap \Omega_{j}$      & samples in node $j$       \\
$D_{t}$ & samples in iteration $t$         &$v_{i}$  & the multiple evaluation metrics of $a_{i}$   & $H_{j}$                            & Hypervolume of $D_{t} \cap \Omega_{j}$ \\ \bottomrule
\end{tabular}
}

\label{notation_definition}
\end{table*}

\begin{figure*}[t]
  \centering
  \includegraphics[width=1\columnwidth]{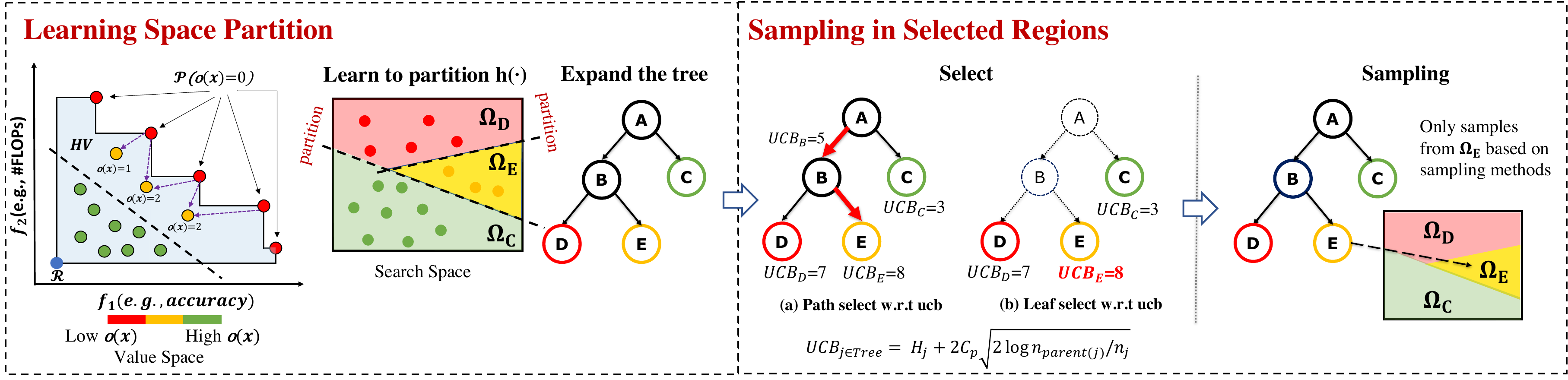}
  \caption{
  \textbf{The overview of a \ours iteration.} 
  The left portion depicts the \emph{Learning Space Partition} phase for optimizing two objectives. The first figure depicts the value space $(f_1, f_2)$ and visualizes the hypervolume $HV$ (blue-shaded area) given the Pareto frontier $P$ and the reference point $R$. The middle figure shows the search space $\Omega$ and its partitions (i.e., $\Omega_{C}$, $\Omega_{D}$, and $\Omega_{E}$) based on samples collected from the previous iterations and their dominance numbers in the objective space. The right figure shows the tree constructed based on the partitions. 
  The right portion depicts the \emph{Sampling in Selected Regions} phase. The left figure visualizes two selection strategies described in \S\ref{sec:monte_carlo_tree_search}. The right figure shows that new architectures will be sampled from the good partition $\Omega_{E}$ with any sampling algorithms.
  Figure adapted from our prior work~\cite{lamoo}.
  }
  \label{fig:workflow}
\end{figure*}

In this section, we present the key details of the learning space partitions based on our previously proposed multi-objective algorithm, referred to as \ours~\cite{lamoo}. 
Briefly, \ours is a \emph{meta-optimization algorithm} that separates good regions out from the entire search space by using observed data. Different multi-objective search algorithms, such as qEHVI~\cite{qehvi}, CMA-ES~\cite{cmaes_moo}, and random sampling, can be combined with \ours in these promising regions for sampling. In this paper, we also introduce a novel promising region selection method, i.e., leaf selection in \S\ref{subsec:search_promising_region_selection}. Table~\ref{notation_definition} lists notations that are used throughout the paper.

Similar to our prior work, LaNAS~\cite{lanas} and LaMCTS~\cite{lamcts}, an iteration of 
 \ours consists of the learning and sampling phases. 
 \ours iterates between learning space partition (\S\ref{sec:learning_phase}) and Monte Carlo Tree Search (\S\ref{sec:monte_carlo_tree_search}) until depleting the sample budget $T$, which can be either search time or the number of samples. Figure~\ref{fig:workflow} presents an overview of an iteration in \oursnas. The pseudo-code is presented in Algorithm~\ref{alg:lamoo}.

\subsection{Learning Partitions in Multi-Objective Search Space}
\label{sec:learning_phase}

\input{algorithm}

\subsubsection{Overview of the Partition Learning Algorithm}

\begin{figure*}[t]
  \centering
  \includegraphics[width=0.8\columnwidth]{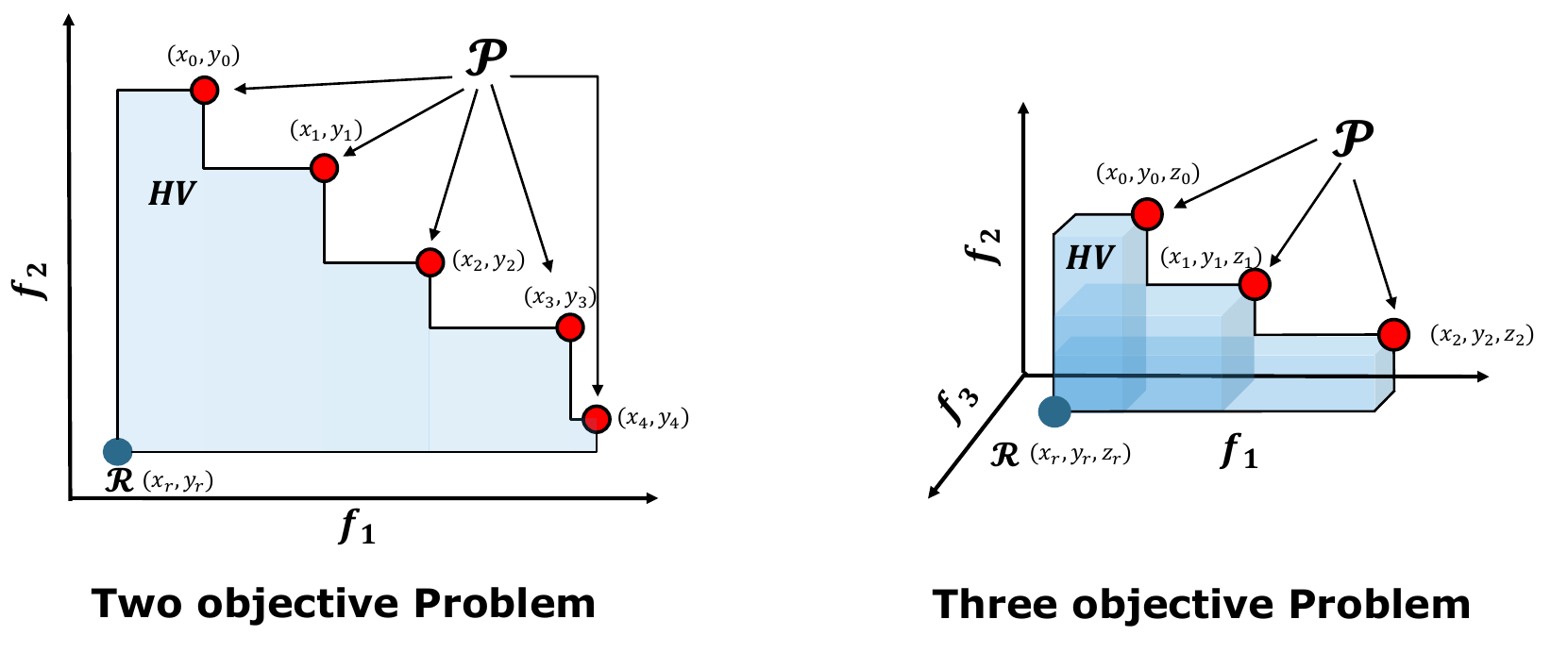}
  \caption{
  Visualization of example hypervolume calculation.
  }
  \label{fig:hv_vis}
\end{figure*}

The learning phase of iteration $t$ begins with $D_{t}$, a dataset consisting of tuples {($a_i$, $v_i$)}, which represent previously observed samples. Here, $a_i$ symbolizes a single architecture represented by an encoding vector. Note that different search spaces have different encoding representations (e.g., variations in vector length and range). Moreover, $v_i$ represents the evaluation metrics for the corresponding architecture $a_i$. Typically, for multi-objective NAS, $v_i$ can include the architecture performance metrics (e.g., accuracy) and inference-phase metrics (e.g., \#params, \#FLOPs, and latency).

For the first iteration, we initialize $D_{t}$ with a few randomly sampled architectures. 
The performance metrics $v_i$ can be obtained in the following three main ways with different data qualities and acquisition costs:
\1 actual training results~\cite{nasnet, alphax, rea}; 
\2 pre-trained dataset results, e.g., NasBench201~\cite{nasbench201, nasbench101}; 
\3  supernet-based estimated results~\cite{few-shot, BG_understanding}. We describe how \oursnas leverages different $v_i$ representations to search multi-objective architectures in \S\ref{subsec:integration_diff_nas}.

During any iteration of the search process, our objective is to identify promising regions (for example, the red region in Figure~\ref{fig:motivation}) from the search space $\Omega$ and then concentrate the search action on these promising regions. To achieve this, we construct a search tree that recursively partitions the search space into \emph{good} and \emph{bad} regions, thereby learning the partitions and identifying the optimal region for the search.

At iteration $t$, with observed samples $D_{t}$ and a root node representing the entire search space $\Omega_{root}$ ($\Omega_{root}$ = $\Omega$), we recursively partition the search space from the root node to the leaves. Specifically, at node $j$, we partition the current search space $\Omega_{j}$ into two disjoint regions, $\Omega_{good}$ and $\Omega_{bad}$, where $\Omega_{j}$ = $\Omega_{good} \cup \Omega_{bad}$. The $\Omega_{good}$ and $\Omega_{bad}$ regions are partitioned based on the rank of the \emph{dominance number} of $D_{t}$. The implementation details are provided in \S\ref{sec:partition}. We quantify the \emph{goodness} of a search space using the metric \emph{hypervolume} (HV) of samples in the space. A larger hypervolume value signifies a more promising space. The definition of HV is provided below.

Given a reference point $R \in \rr^M$ (e.g., as shown in Figure~\ref{fig:hv_vis}), the \emph{hypervolume} of a finite approximate Pareto set $\mathcal{P}$ is the M-dimensional Lebesgue measure $\lambda_{M}$ of the space dominated by $\mathcal{P}$ and bounded from below by $R$. That is, $HV(\mathcal{P}, R) = \lambda_{M} (\cup_{i=1}^{|\mathcal{P}|}[R, y_{i}])$, where $[R, y_{i}]$ denotes the hyper-rectangle bounded by reference point $R$ and $y_{i}$. 
We present visualizations for two and three-objective optimization examples in Figure~\ref{fig:hv_vis}. The hypervolume is indicated by the blue-shaded area. For a two-objective problem, the hypervolume $HV(\mathcal{P}, R)$ is calculated as $\sum_{i=1}^{n-1} (x_{i} - x_{i-1}) \cdot (y_{i} - y_{r}) +  (x_{0} - x_{r}) \cdot (y_{0} - y_{r})$. The hypervolume computation for a three-objective problem is more complex. Considering a simplified scenario where $(x_{0} < x_{1} < x_{2}) \land (y_{0} > y_{1} > y_{2}) \land (z_{0} > z_{1} > z_{2})$, the hypervolume $HV(\mathcal{P}, R)$ is given by $(x_{0} - x_{r}) \cdot (y_{0} - y_{r}) \cdot (z_{0} - z_{r}) + (x_{1} - x_{0}) \cdot (y_{1} - y_{r}) \cdot (z_{1} - z_{r}) + (x_{2} - x_{1}) \cdot (y_{2} - y_{r}) \cdot (z_{2} - z_{r})$.

The resulting partitions then satisfy the following property of $H_{good}$ > $H_{bad}$, where $H_{good}$ is the hypervolume calculated based on $D_{t} \cap \Omega_{good}$ and $H_{bad}$ is based on $D_{t} \cap \Omega_{bad}$.
By repeating the partitioning process, we can construct a tree comprising nodes that partition the entire search space into diverse performance regions, in terms of hypervolume, for multi-objective optimization. The following illustrates the node-level partitioning in detail.

\subsubsection{Details of Node-Level Partitioning}
\label{sec:partition}

At each node $j$, for each sampled architecture $\va \in $ $D_{t} \cap \Omega_{j}$, we calculate its \emph{dominance number} $o_{t,j}(\va)$, defined in \eqref{eq:dominance}, to represent its performance. To speed up the calculation, we use Maxima Set~\citep{maximaset} which runs in $O(|D_{t,j}|\log |D_{t,j}|)$, compared to naive computation which requires $O(|D_{t,j}|^2)$ operations. Note that architectures in the Pareto frontier have a dominance number of 0. In other words, the lower the dominance number, the better the architecture. We, therefore, sort all sampled architectures ($D_{t} \cap \Omega_{j}$) based on their dominance numbers in descending order and label the first half as good samples and the remaining as bad samples.

After all samples are labeled in node $j$, we construct a SVM classifier $h(\cdot)$ fitting the labeled architectures $c(\va)$ as below: 

\begin{equation}
\mathop {min}\limits_{a_{i} \in D_{t} \cap \Omega_{j}} \sum_{i} (h(a_{i}) \oplus c(a_{i})).
\label{eq:cls_min}
\end{equation}

Given that we have categorized all sampled architectures ($D_{t} \cap \Omega_{j}$) as either good or bad, we formulate the task of partitioning the space as a binary classification problem. In this scenario, $h(a_{i}), c(a_{i}) \in \{0, 1\}$, with $\oplus$ symbolizing the XOR operation. Within this framework, 0 signifies a bad architecture, while 1 indicates a good one. This implies that $c(a_{i}) = 1$ denotes a good architecture, while $c(a_{i}) = 0$ indicates a bad architecture. \Eqref{eq:cls_min} is equivalent to $h(a_{i})\cdot(1 - c(a_{i})) + c(a_{i})\cdot(1 - h(a_{i}))$. Consequently, the minimum of the above equation is 0 if all samples are classified correctly. In the worst-case scenario, where all samples are misclassified, the value equates to the number of samples.

Upon completion of the classifier training, the search space $\Omega_{J}$ bifurcates to a good and a bad region (i.e., $\Omega_{good}$ and $\Omega_{bad}$) by the $h(\cdot)$. For ease of exposition, we designate the left child of node $j$ as the good region and the right child as the bad region of $\Omega_{j}$.

The search space can be continuously partitioned, and the corresponding search tree can be constructed by repeating the aforementioned steps until one of the stopping conditions is met. We consider the following three conditions:
\1 If the samples in a node cannot be split by $h(\cdot)$, this node will be marked as a leaf node. Here, a node is considered non-splittable if all samples receive the same label or if the classifier cannot classify based on the current samples (i.e., always predict to 0 or 1). \2 If the tree reaches the maximum height, which is considered a hyper-parameter in this work, the process will halt. \3 If the number of samples in a node is less than the \emph{minimal sample threshold}, which is also a hyper-parameter, this node will be marked as a leaf node.

Once the entire space is partitioned (i.e., the search tree is constructed), partitions represented by leaf nodes follow the sequence $H_{leftmost}$ > \dots > $H_{rightmost}$, with the leftmost leaf node representing the most promising partition. A detailed visualization of the space partitioning and its effectiveness is shown in Figure~\ref{fig:verf}.

\subsection{Sampling from the Promising Region with MCTS}
\label{sec:monte_carlo_tree_search}

Once the space partitions are learned as previously described, the next step is to search the constructed tree. The primary goal is to sample the most promising neural architectures from the selected regions and use the obtained $(\va_{i},\vv_{i})$ tuples to update the learning phase for the next iteration. We employ the Monte Carlo Tree Search (MCTS) algorithm to explore and exploit the learned partitions, thereby preventing us from getting stuck in local optima~\cite{lanas}. Similar to the traditional MCTS algorithm, our LaMOO search incorporates \emph{selection}, \emph{sampling}, and \emph{backpropagation} stages. We omit the expansion part of MCTS because our search tree, including the structure and the learned classifiers, will remain fixed during the search phase. We elaborate on two strategies for selecting promising regions in \S\ref{subsec:search_promising_region_selection} and suitable sampling methods in \S\ref{subsec:search_sampling_from_a_leaf}.

\subsubsection{Promising Region Selection Strategies}
\label{subsec:search_promising_region_selection}

In this work, we consider two selection strategies for implementing the MCTS. The \emph{path selection} strategy derives from the original MCTS algorithm, while the \emph{leaf selection} strategy is a computation-efficient variation that saves Hypervolume calculation. Figure~\ref{fig:verf} visualizes the resultant partitions and Pareto frontier with the path selection strategy.

\emph{\textbf{Path selection}} works by traversing down the constructed tree to generate the most promising path. The search starts from the root node and stops when a leaf node is reached. At a given node, the agent determines which child node to traverse based on the UCB1 value~\cite{ucb}. The UCB1 value of node $j$ is defined as:
\begin{equation}
\mathrm{UCB}_{j} := H_{j} + 2C \sqrt{\frac{2\log n_{\mathrm{parent(j)}}}{n_{j}}}.
\label{eq:ucb}
\end{equation}

In this equation, $n_{j}$ denotes the number of samples in node $j$, $n_{\mathrm{parent(j)}}$ refers to the number of samples in node $j$'s parent, $C$ is a tunable hyperparameter that adjusts the degree of exploration, and $H_{j}$ denotes the hypervolume of the samples in node $j$. The first term $H_{j}$ is the exploitation term which evaluates the expected multi-objectives performance of samples in the current node $j$. The second term, $2C \sqrt{\frac{2\log n_{\mathrm{parent(j)}}}{n_{j}}}$, represents the exploration term that encourages the selection of the next node with fewer samples. To summarize, given two sibling nodes $i$ and $j$, we will traverse node $i$ if $UCB_i \geq UCB_j$. Hence, the path selection strategy ensures all nodes in the chosen path possess a larger UCB value compared to their sibling node. However, this search strategy demands the computation of the hypervolume for every node, which can be highly resource-intensive. Prior work~\cite{hv_comp1} shows that when the number of objectives $M$ is more than three, the computation cost of hypervolume is $O(N^{\frac{M}{2}} + N\log{N})$, where $N$ is the number of searched samples in total\footnote{When $M \leq 3$, the computation complexity of hypervolume is $O(N\log{N})$.}. That is, the hypervolume computation cost is growing exponentially with $M$ when $M>3$. Next, we describe the \emph{leaf selection} strategy, which mitigates this high computation cost problem.

\emph{\textbf{Leaf selection}} significantly reduces the hypervolume computation by only calculating the UCB1 value for all the leaf nodes. The leaf node with the highest UCB1 value is selected as the promising region. This strategy avoids computing the hypervolume in the non-leaf nodes of the tree, where hypervolume calculation is the primary computational cost of \oursnas, especially in many-objective problems (i.e., $M > 3$). Moreover, even if the number of objectives is less than three, our leaf selection strategy can still save computational time compared to the path selection strategy. This is because, in this case, the number of samples becomes the dominant factor in hypervolume computation. Generally, the number of samples in the leaf nodes is fewer than in the non-leaf nodes, especially nodes nearer to the root. Figure~\ref{fig:ablation_d} shows that the node selection strategy can achieve a similar sample performance with less time compared to the path selection strategy.

\subsubsection{Sampling from The Promising Leaf Node}
\label{subsec:search_sampling_from_a_leaf}

\begin{figure}[t]
\centering 
\subfloat[][NasBench201]{\includegraphics[width=.32\textwidth]{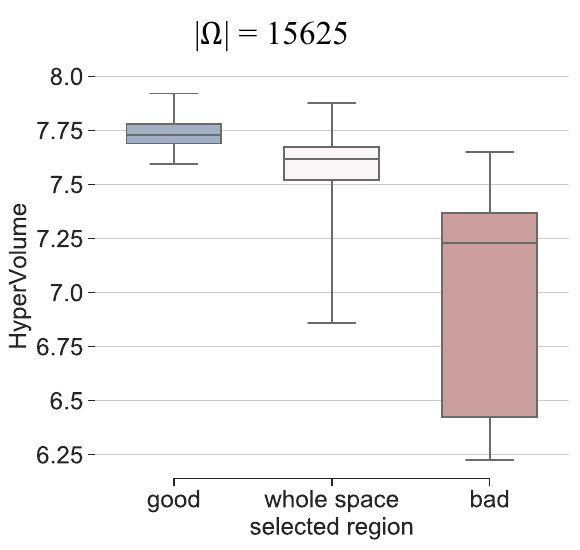}\label{fig:region_sample_a}} \quad
\subfloat[][NasBench201 with supernet]{\includegraphics[width=.31\textwidth]{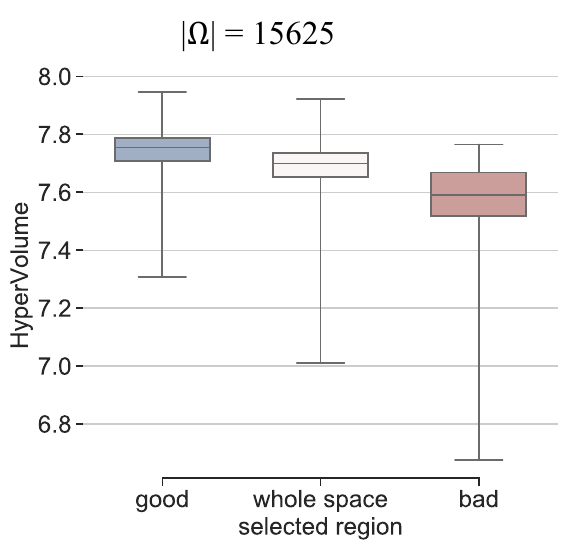}\label{fig:region_sample_b}}  \quad 
\subfloat[][NasBench301]{\includegraphics[width=.31\textwidth]{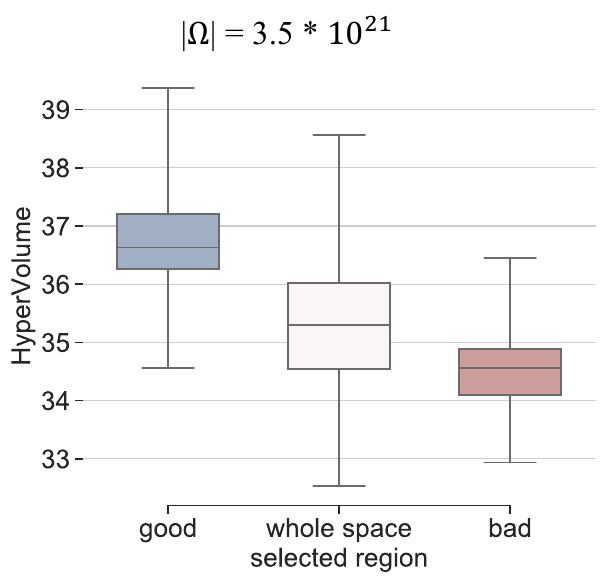}\label{fig:region_sample_c}}  \quad \\
\caption{The range of hypervolume for 50 samples randomly generated from different regions in NasBench201, NasBench201 by supernet search, and NasBench301.}
\label{fig:region_sample}
\end{figure}

After selecting the promising leaf node $j$, we can utilize multi-objective sampling methods such as random sampling, qEHVI~\citep{qehvi}, and CMA-ES~\citep{cma-es} to select samples. Therefore, we regard \ours as a \emph{meta-algorithm} that can enhance the search performance of existing MOO algorithms.

\emph{\textbf{Random sampling.}} We implement the \emph{reject sampling} technique in which we randomly sample the promising partition $\Omega_{j}$ and only keep samples that satisfy the constraint. Here, we consider each $h(\cdot)$ in the nodes as a constraint and only select samples that are classified as good in the selected path. 
To understand the effectiveness of \ours even with random sampling, we conduct a preliminary evaluation as follows.
Specifically, we randomly generate ten samples from NasBench201 or NasBench301 as the initialization. In each iteration, we update 5 batched samples for all search algorithms and 300 samples in total. The setup of \ours followed ~\cite{lamoo}. We run \ours 5 times and run the random sampling for each region 150 times. 
We also evaluate the effectiveness of learning partitions in one-shot NAS on Nasbench201, labeled as \emph{Nasbench201 with supernet}. We train a supernet based on ~\cite{few-shot} and leverage the supernet to estimate the accuracy of sampled architecture. All the search and sampling processes are based on the results from the supernet. Figure~\ref{fig:region_sample} compares the random sampling performance where each box plot denotes the hypervolume distribution with randomly generated 50 samples in different selected regions, i.e., good, whole space, and bad.  
We observe that searching in the good region for NasBench201 improves the median hypervolume and leads to a tighter distribution compared to searching in the entire space, as shown in Figure~\ref{fig:region_sample_a}. 
Similar trends are observed for NasBench301 as shown in Figure~\ref{fig:region_sample_c}.
Moreover, we see that compared to standard NAS, one-shot NAS (Figure~\ref{fig:region_sample_b} has a worse hypervolume distribution in the good region but better results in the bad region\footnote{Note that the hypervolume calculation is based on true accuracy in NasBench201.}. This is largely in part because the estimation by the supernet is not accurate as claimed by prior work~\cite{BG_understanding}, which results in inaccurately learned partitions as well.

\emph{\textbf{Bayesian optimization.}} A typical Bayesian Optimization for NAS works by first training a surrogate model using a Gaussian Process Regressor (GPR) on observed architectures and then generating new architectures based on the acquisition function, such as Expected Improvement (EI) or Upper Confidence Bound (UCB)~\cite{ei, ucb}. In this work, we integrate \ours with a state-of-the-art multi-objective Bayesian Optimization solver named qEHVI~\citep{qehvi}, which finds samples to optimize parallel version of the acquisition function called Expected Hypervolume Improvement (EHVI). 
To incorporate qEHVI into \ours's sampling step, we confine qEHVI’s search space to a sub-space represented as a node in the MCTS tree. As claimed in our previous space partition work ~\cite{lamoo}, \ours leverages previous samples to learn the partitions, which convert complicated non-convex optimization of the acquisition functions in Bayesian optimization into a simple traversal of hierarchical partition tree while still precisely capturing the promising regions for the sample proposal, reducing the complexity of Bayesian Optimization.

\emph{\textbf{Evolutionary algorithms.}} A typical evolutionary algorithm (EA) search step consists of two parts. The first part is \emph{selection}, which chooses several samples with the best performance. The second part is \emph{mutation}, which slightly alters the selected samples to propose new individuals~\cite{cma-es, rea}. Take the NAS task as an example. The mutation part changes 1-2 connections or existing operations of the selected best architectures to evolve the entire population in the current step~\cite{rea}. In this work, we combine \oursnas with CMA-ES~\citep{cma-es}, an evolutionary algorithm originally designed for multi-objective optimization. To apply CMA-ES in the multi-objective NAS, we use CMA-ES to sample the promising region $\Omega_{j}$ by picking an architecture that minimizes the dominance number $o(\vx)$.
Note that $o(\vx)$ changes over iterations, so we need to update $o(\vx)$ with new samples before running CMA-ES.

\subsubsection{Optional Backpropagate Rewards}

A distinctive trait of MCTS is giving more preference to exploration by backpropagation. Backpropagation backtracks the selected path from the leaf to the root, updating the visit counts and values of the samples. Specifically, for node $j$ on the selected path, the visit count $n_{j}$ is updated with the number of new samples, and so is the hypervolume $H_{j}$. This backpropagation step assists in evolving our search tree from one iteration to the next. However, this step can be bypassed by directly reconstructing the tree using the newly generated samples. We use this non-backpropagation version in our experiments throughout the paper.

\section{Multi-Objective Learning Space Partitions with Different NAS Methods}
\label{subsec:integration_diff_nas}

\begin{table*}[t]
\centering
 \makeatletter\def\@captype{table}\makeatother\caption{Comparisons of existing NAS methods.}
 
\label{tab:nas_comparison}
\resizebox{\textwidth}{!}{%
\begin{threeparttable}
\begin{tabular}{@{}cccccc@{}}
\toprule
\textbf{NAS method}                      & \textbf{Search method}                        & \textbf{One(few)-shot support}                    & \textbf{Performance prediction}       & \textbf{Multi-objective}              & \textbf{Meta-optimizer}               \\ \midrule
\multicolumn{1}{|c|}{NASNet~\cite{nasnet}}    & \multicolumn{1}{c|}{Reinforcement learning}              & \multicolumn{1}{c|}{}        & \multicolumn{1}{c|}{}        & \multicolumn{1}{c|}{}        & \multicolumn{1}{c|}{}        \\
\multicolumn{1}{|c|}{ENAS~\cite{enas}}      & \multicolumn{1}{c|}{Reinforcement learning}              & \multicolumn{1}{c|}{$\surd$} & \multicolumn{1}{c|}{}        & \multicolumn{1}{c|}{}        & \multicolumn{1}{c|}{}        \\
\multicolumn{1}{|c|}{MnasNet~\cite{mnasnet}}   & \multicolumn{1}{c|}{Reinforcement learning}              & \multicolumn{1}{c|}{}        & \multicolumn{1}{c|}{}        & \multicolumn{1}{c|}{$\surd$\tnote{$\dagger$}} & \multicolumn{1}{c|}{}        \\ \midrule
\multicolumn{1}{|c|}{AmoebaNet~\cite{rea}} & \multicolumn{1}{c|}{Evolutionary algorithm}              & \multicolumn{1}{c|}{}        & \multicolumn{1}{c|}{}        & \multicolumn{1}{c|}{}        & \multicolumn{1}{c|}{}        \\
\multicolumn{1}{|c|}{LEMONADE~\cite{lamonade}}  & \multicolumn{1}{c|}{Evolutionary algorithm}              & \multicolumn{1}{c|}{}        & \multicolumn{1}{c|}{$\surd$} & \multicolumn{1}{c|}{$\surd$} & \multicolumn{1}{c|}{}        \\
\multicolumn{1}{|c|}{NSGANetV2~\cite{nsganetv2}} & \multicolumn{1}{c|}{Evolutionary algorithm}              & \multicolumn{1}{c|}{$\surd$} & \multicolumn{1}{c|}{$\surd$} & \multicolumn{1}{c|}{$\surd$} & \multicolumn{1}{c|}{}        \\
\multicolumn{1}{|c|}{ChamNet~\cite{chamnet}}   & \multicolumn{1}{c|}{Evolutionary algorithm}              & \multicolumn{1}{c|}{$\surd$} & \multicolumn{1}{c|}{}        & \multicolumn{1}{c|}{$\surd$\tnote{$\dagger$}} & \multicolumn{1}{c|}{}        \\
\multicolumn{1}{|c|}{OFANet~\cite{ofa}}    & \multicolumn{1}{c|}{Evolutionary algorithm}              & \multicolumn{1}{c|}{$\surd$} & \multicolumn{1}{c|}{}        & \multicolumn{1}{c|}{$\surd$\tnote{$\dagger$}} & \multicolumn{1}{c|}{}        \\ \midrule
\multicolumn{1}{|c|}{PNAS~\cite{pnas}}      & \multicolumn{1}{c|}{SMBO\tnote{$\ddagger$}}            & \multicolumn{1}{c|}{}        & \multicolumn{1}{c|}{$\surd$} & \multicolumn{1}{c|}{}        & \multicolumn{1}{c|}{}        \\ 
\multicolumn{1}{|c|}{DPP-Net~\cite{dppnet}}      & \multicolumn{1}{c|}{SMBO\tnote{$\ddagger$}}            & \multicolumn{1}{c|}{}        & \multicolumn{1}{c|}{$\surd$} & \multicolumn{1}{c|}{$\surd$\tnote{$\dagger$}}        & \multicolumn{1}{c|}{}        \\
\midrule
\multicolumn{1}{|c|}{DARTS~\cite{DARTS}}     & \multicolumn{1}{c|}{Gradient}        & \multicolumn{1}{c|}{$\surd$} & \multicolumn{1}{c|}{}        & \multicolumn{1}{c|}{}        & \multicolumn{1}{c|}{}        \\
\multicolumn{1}{|c|}{PCDARTS~\cite{pcdarts}}   & \multicolumn{1}{c|}{Gradient}        & \multicolumn{1}{c|}{$\surd$} & \multicolumn{1}{c|}{}        & \multicolumn{1}{c|}{}        & \multicolumn{1}{c|}{}        \\ 
\multicolumn{1}{|c|}{FBNetV2~\cite{fbnetv2}}   & \multicolumn{1}{c|}{Gradient}        & \multicolumn{1}{c|}{$\surd$} & \multicolumn{1}{c|}{}        & \multicolumn{1}{c|}{$\surd$\tnote{$\dagger$}}        & \multicolumn{1}{c|}{}        \\ \midrule
\multicolumn{1}{|c|}{LaNAS~\cite{lanas}}     & \multicolumn{1}{c|}{Space partition} & \multicolumn{1}{c|}{$\surd$} & \multicolumn{1}{c|}{} & \multicolumn{1}{c|}{}        & \multicolumn{1}{c|}{$\surd$} \\ 
\multicolumn{1}{|c|}{\textbf{LaMOO (ours)}}     & \multicolumn{1}{c|}{Space partition} & \multicolumn{1}{c|}{$\surd$} & \multicolumn{1}{c|}{$\surd$} & \multicolumn{1}{c|}{$\surd$} & \multicolumn{1}{c|}{$\surd$} \\ \bottomrule
\end{tabular}%
\begin{tablenotes}
    \item[$\dagger$] Optimization leverages scalarized multiple objectives (e.g., objective in Figure~\ref{fig:two_c} or additional constrains (e.g., objective in Figure~\ref{fig:two_a}\&Figure~\ref{fig:two_b}). 
    \item[$\ddagger$] Sequential Model Based Optimizations.
  \end{tablenotes}
\end{threeparttable}
}
\end{table*}

\begin{figure}[t]
\centering 
\includegraphics[width=0.8\columnwidth]{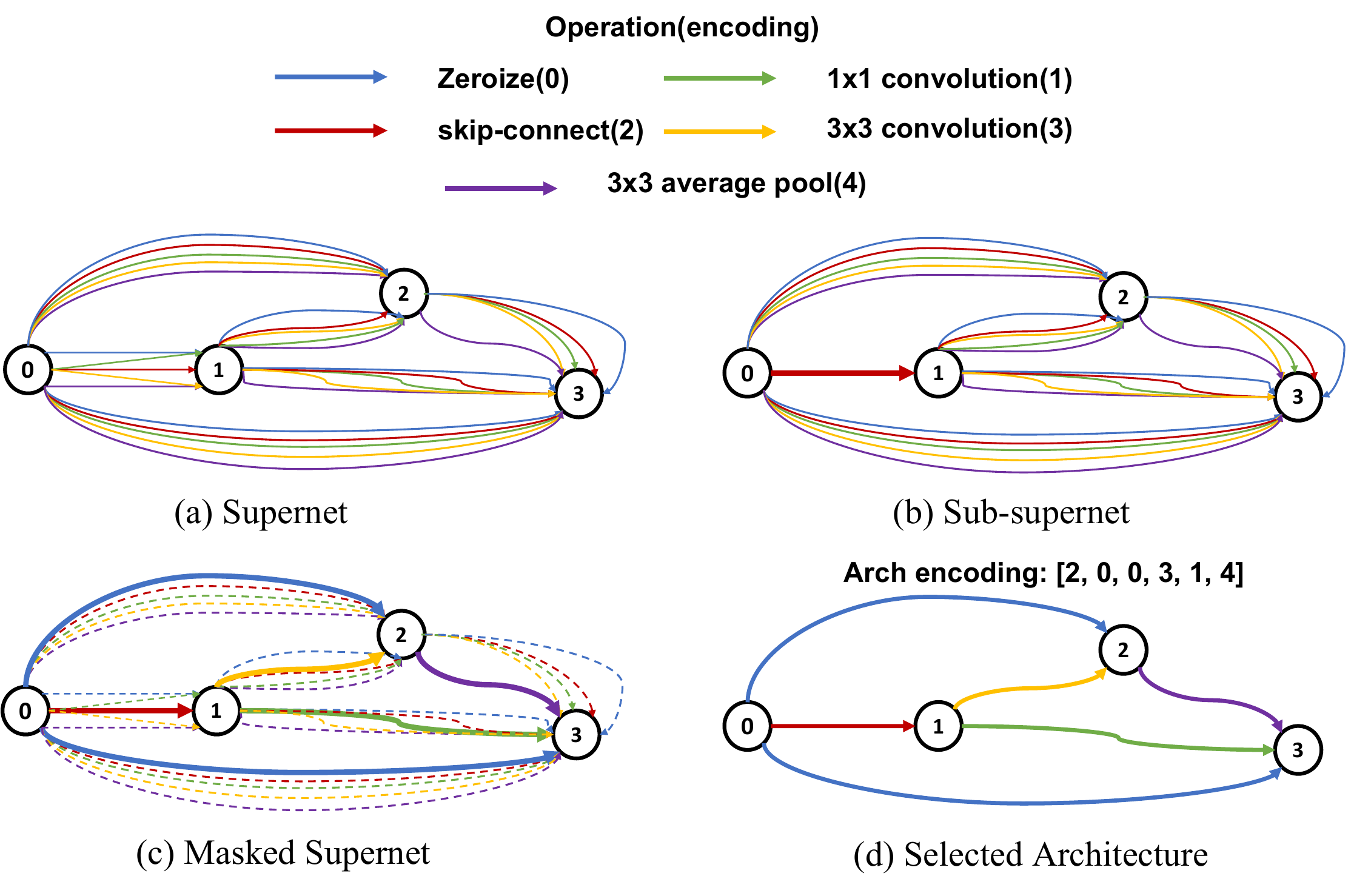}
\caption{
The cell structure of the supernet used in Nasbench201. (a) is a supernet structure of the NasBench201 cell. Any edge between two nodes is combined with four operations. (b) is a sub-supernet of the NasBench201 cell. The edge between node 0 and node 1 has only one operation. (c) demonstrates a masked supernet that only one path is activated. The dotted lines are inactive edges. (d) is a cell architecture example from NasBench201 with its encoding.  
}
\label{fig:nasbench201_arch}
\end{figure}

In this section, our goal is to demonstrate the efficacy of \ours when applied to the problem of \emph{multi-objective neural architecture search}. We detail how \ours can be combined with three major NAS evaluation approaches to enhance search efficiency. These approaches differ in how we obtain the architecture and its performance metrics ($a_i$, $v_i$), and the associated evaluation costs. Table~\ref{tab:nas_comparison} compares popular NAS algorithms from multiple perspectives. Our \ours is the only NAS method supporting all distinct NAS evaluation approaches (one-shot~\cite{BG_understanding,yu2019evaluating,few-shot, enas}, few-shot~\cite{few-shot}, and performance prediction~\cite{pnas}); moreover, \ours is compatible with multi-objective optimization and can act as a meta-optimizer to enhance other NAS methods.

\subsection{One-Shot Neural Architecture Search}
\label{sec:one-shot}

Prior work proposed to mitigate the high evaluation cost in vanilla NAS~\cite{nasnet, rea, alphax} with a weight-sharing technique, avoiding retraining sampled networks from scratch~\cite{enas}. This type of NAS is referred to as \emph{one-shot neural architecture search}. Typically, one-shot NAS first trains an over-parameterized \emph{supernet} that covers all potential operations and connections of the entire search space and searches on the supernet as an integrated bi-level optimizations~\cite{DARTS, pcdarts}. Figure~\ref{fig:nasbench201_arch}(a) shows an example of the supernet topology used for the NasBench201 search space.

Besides the bi-level optimization approach~\cite{DARTS}, another variant of one-shot NAS uses the supernet as a performance estimator to predict the architecture's performance~\cite{BG_understanding,yu2019evaluating,few-shot, enas}. The performance of any architecture from the search space can be estimated using a well-trained supernet by sharing common weights.

Our \oursnas integrates with the latter variant of one-shot NAS, which efficiently assesses the performance of sampled architectures (e.g., accuracy). Note that due to co-adaptation among operations~\cite{BG_understanding}, the performance assessed by the supernet is not as accurate as the performance obtained through actual training of the sampled architectures. \S\ref{subsec:few_shot_nas} elaborates on few-shot NAS~\cite{few-shot}, which enhances the evaluation performance. The following describes the supernet design, its training, and its usage in detail.

\emph{\textbf{Supernet design.}} We use the following three supernet designs in this work. 
\begin{itemize}
    \item \emph{NasBench201}~\cite{nasbench201}. 
    Our supernet design follows \cite{how_to_train, few-shot}. Figure~\ref{fig:nasbench201_arch}(a) shows the topology of the supernet in the NasBench201 search space. The supernet keeps the same nodes as the architectures in the search space. However, each node in the supernet is connected by all possible operations with weight addition. 
    Figure~\ref{fig:nasbench201_arch}(d) illustrates an architecture that includes four feature map nodes and six operation edges. 
    \item \emph{DARTS search space on CIFAR10}~\cite{DARTS} . On the CIFAR10 classification task, we use the same supernet design as DARTS~\cite{DARTS}. 
    The network architecture is built by stacking multiple normal cells and reduction
    cells. We only perform NAS on the structure inside of the cells. 
    The normal cell keeps the same dimensions as the input feature map, while the reduction cell cuts the height and width of the feature map by two and multiplies the channel numbers by two.
    The supernet consists of four nodes, which are connected by all available candidate operations.
    We replace the bi-optimization search process on DARTS with the random mask training because we only use the supernet as a performance estimator for \oursnas. 
    \item \emph{EfficientNet search space on ImageNet}~\cite{efficientnet}. On the ImageNet classification task, we follow the supernet design by \cite{ofa}. 
    The once-for-all supernet supports predicting different architectures with different depths, widths, kernel sizes, and resolutions without retraining. 
   
\end{itemize}

\emph{\textbf{Supernet training.}} In this work, we leverage the random mask strategy~\cite{uniform_random_train} to train a supernet to convergence. This is because this training strategy was shown to enhance the supernet evaluation performance regarding rank co-relation~\cite{uniform_random_train, few-shot}. Specifically, instead of training the entire supernet in each iteration, we randomly pick one architecture from the search space, train it for one epoch, and update the corresponding weights in the supernet.

\emph{\textbf{Architecture evaluation by supernet.}} After training the supernet, any architecture from the search space can be evaluated by the supernet. For evaluating a specific architecture like the one shown in Figure~\ref{fig:nasbench201_arch}(d), we first mask out all other unused edges (dotted line as shown in Figure~\ref{fig:nasbench201_arch}(c)), and use the supernet with the remaining weights to estimate the architecture performance.

\subsection{Few-shot Neural Architecture Search}
\label{subsec:few_shot_nas}

While one-shot supernet enables quick estimation of architectures in the search space, many works~\cite{yu2019evaluating, few-shot,nao,nasbench201,renqianBalance,BG_understanding} show that supernet often leads to search performance degradation due to inaccurate architecture performance prediction. To address this downside of supernet, \cite{few-shot} proposed few-shot NAS which greatly reduces the negative impact of co-adaptation among operations. Specifically, instead of using one supernet, the search space is represented with multiple sub-supernet, each covering a part of the search space. Each sub-supernet, therefore, uses fewer operations in a compound edge. Figure~\ref{fig:nasbench201_arch}(b) is an example of a sub-supernet from the same search space. The compound edge between node 0 and node 1 is simplified to one operation. 

In short, because prior work demonstrates that few-shot NAS can greatly improve the search performance compared to the one-shot NAS~\cite{few-shot, few-shot_gradient, fewshottaskagnostic, naslid, kshotnas}, we also integrate \oursnas with few-shot NAS in this work. 
The training strategy of few-shot NAS is similar to that of one-shot NAS. To speed up the training procedure of sub-supernets, few-shot NAS leverages the weight-transferring technique. Specifically, a supernet is trained from scratch until convergence. All sub-supernet training then directly leverage the weights from the supernet and only take a few epochs to converge. This training strategy can save much training time compared to training all sub-supernets from scratch.

The architecture evaluation is similar to the steps described above for one-shot NAS. We first pick the corresponding sub-supernet and mask the unused operations in the sub-supernet to estimate the architectures' performance.

\subsection{Performance Predictor Guided Search}

Besides supernet, another efficient NAS approach is to train performance predictors (e.g., deep neural network) based on observed architectures to guide the search process~\cite{pnas, alphax, dppnet}. The accuracy of such performance predictors, in part, depends on the information about architectures, e.g., the number of trained architectures. In this work, we integrate the performance predictor from NasBench301, which was demonstrated to have good prediction accuracy~\cite{nasbench301}, with \oursnas. This predictor is trained on 60K architectures.

\section{Experiments}
We evaluate our \oursnas algorithm on two types of NAS scenarios. The first type is based on three popular NAS datasets, NasBench201~\cite{nasbench201}, NasBench301~\cite{nasbench301} and HW-NAS-Bench~\cite{hwnasbench}. The second type is real-world deep learning domain applications, including image classification, object detection, and language models.

\subsection{NAS Datasets}
To date, there are three popular open-source NAS datasets, NasBench101~\cite{nasbench101}, NasBench201~\cite{nasbench201}, and NasBench301~\cite{nasbench301}. 
For our evaluation, we chose the latter two because the network architectures in these two datasets cover the entire search space, while the NasBench101 dataset only consists of a small subset of architectures. 
Evaluating using NasBench101 is challenging because we will not have access to important information, such as accuracy, during the search. 

We also use HW-NAS-Bench~\cite{hwnasbench}, a hardware-aware neural architecture search benchmark, which offers extensive metrics of architectures in NasBench201~\cite{nasbench201} for many-objective NAS.

For each NAS dataset, we evaluate the search performance of using \ours in conjunction with two SoTA algorithms called qEHVI~\cite{qehvi} and CMAES~\cite{cma-es}. 
Specifically, qEHVI and CMA-ES are Bayesian optimization (BO)-based and evolutionary-based multi-objective optimization algorithms, respectively.  The other search algorithms we evaluated include qPAREGO~\cite{qparego} (BO-based), NSGA-family~\cite{nsga-ii, nsgaiii} (evolutionary-based), and MOEAD~\cite{moead} (evolutionary-based).

\subsubsection{Evaluation Using NasBench201}
\label{subsec:nasbench201}

\para{Dataset overview.} 
NasBench201 is an open-source benchmark and dataset for evaluating NAS algorithms~\cite{nasbench201}. 
In Nasbench201, the architectures are formed by stacking the cells together. 
Figure~\ref{fig:nasbench201_arch}(a) depicts the design of a cell as a fully-connected graph. Specifically, each cell contains 4 nodes and 6 edges. Each node is a feature map, and the edge represents a type of operation. A pair of nodes are connected by one of the following operations, i.e., zeroize, skip-connect, 1x1 convolution, 3x3 convolution, and 3x3 average pooling. To represent each architecture uniquely, we encoded the five operations with the numbers 0 to 4 and used a 6-length vector to represent a specific architecture. 

We chose two objectives, \#FLOPs and accuracy, to optimize.
We normalized \#FLOPs to range $[-1, 0]$ and accuracy to $[0, 1]$. 
NasBench201 provides all architectures' information in its search space and comprises 15625 architectures trained to converge on CIFAR10~\cite{cifar10}. 
As such, NAS algorithms can leverage the preexisting information about each architecture's \#FLOPs and accuracy as ground truth to avoid time-consuming training during algorithm evaluation. 
After the normalization, we also calculated the maximal hypervolume according to the ground truth of all architectures.

\para{Metric.}
We used the \emph{log hypervolume} difference, the same as~\cite{qehvi}, as our evaluation criterion for NasBench201. 
This is because, in NasBench201, the performance difference between any two architectures may be small. 
Using log hypervolume allows us to visualize the sample efficiency of different algorithms more effectively.
The metric is defined as:

\begin{equation}
HV_{\mathrm{log\_diff}} := \log(HV_{\mathrm{max}} - HV_{\mathrm{cur}}),
\label{log_hypervolume}
\end{equation}
where $HV_{\mathrm{cur}}$ is the hypervolume of current samples obtained by the algorithm with a given budget.
The smaller the log hypervolume difference, the better the performance.

\begin{figure}[t]
\centering 
\subfloat[][Bayesian optimization]{\includegraphics[width=.48\textwidth]{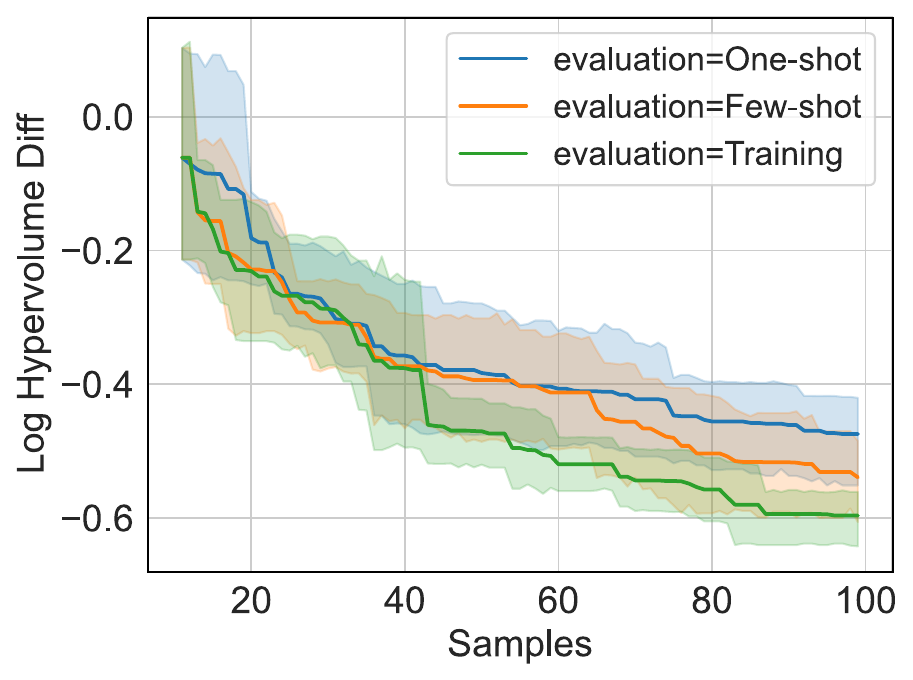}\label{oneshot_baye}} \quad
\subfloat[][Random search]{\includegraphics[width=.48\textwidth]{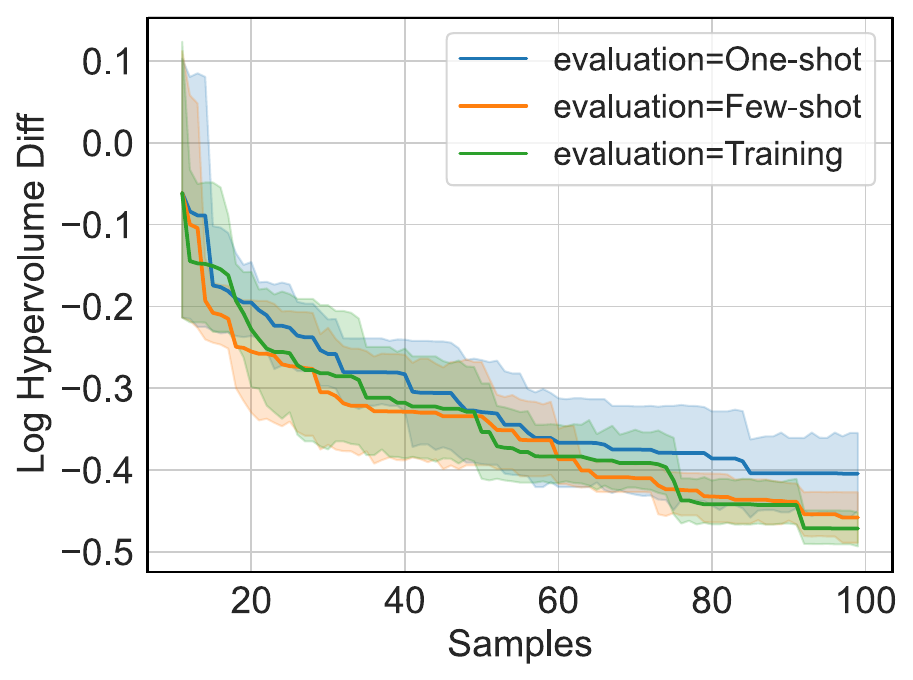}\label{oneshot_random}}  \quad \\
\caption{One-shot, few-shot, and vanilla (i.e., evaluation with  trained architectures) NAS comparison on NasBench201. We ran each algorithm seven times (the shaded areas are $\pm$ std of the mean).}
\label{fig:oneshot_nasbench201}
\end{figure}

\begin{figure}[t]
\centering 
\subfloat[][Bayesian optimization]{\includegraphics[width=.478\textwidth]{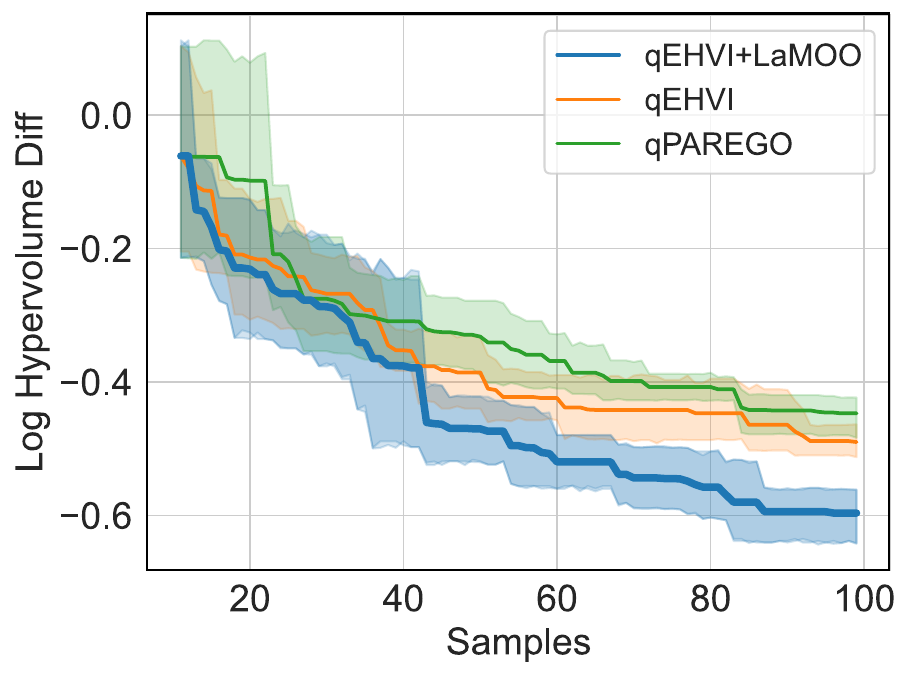}\label{201_baye}} \quad
\subfloat[][Evolutionary search]{\includegraphics[width=.49\textwidth]{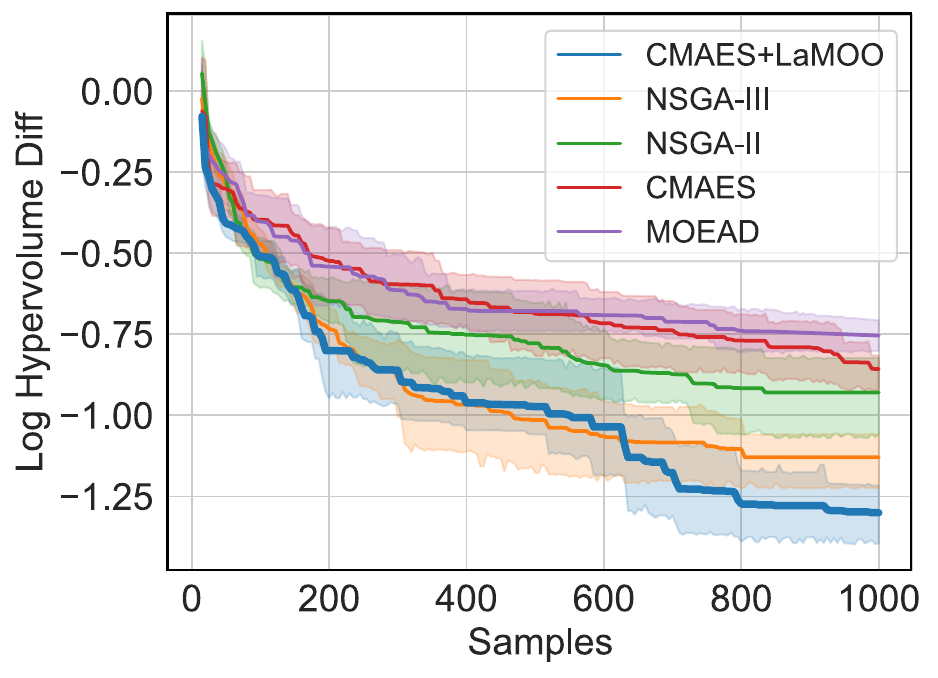}\label{201_evo}}  \quad \\
\caption{The search performance of \ours on NasBench201.  We ran each algorithm seven times (the shaded area are $\pm$ std of the mean).}
\label{fig:nasbench201}
\end{figure}

\para{Results.}
Figure~\ref{fig:oneshot_nasbench201} first compares the search performance of three NAS approaches when integrated with \oursnas. 
One-shot evaluation has the worst search performance due to the inaccurate accuracy estimation in either Bayesian search or random search. The few-shot version is better, and vanilla NAS, by training each architecture from scratch, performs best. Note that the inaccuracy of neural architecture performance estimation by one-shot or few-shot NAS may mislead \ours. Specifically, architectures that appear promising based on supernet estimations but actually have low accuracy may be considered good architectures for classification. This misidentification can lead \ours to focus subsequent sampling efforts in non-promising regions, thus degrading overall search performance as most generated samples may come from the non-promising regions and not effectively guide search directions or space partitioning. However, the exploration component within the MCTS has the potential to fix the search direction by preventing \ours from being confined to these less promising areas.

Next, we evaluate the efficacy of \ours as a meta-algorithm integrated with vanilla NAS. As shown in Figure~\ref{fig:nasbench201}, \oursnas with qEHVI outperforms all our BO baselines, and \oursnas with CMA-ES outperforms all our EA baselines, in terms of $HV_{\mathrm{log\_diff}}$.
Specifically, LaMOO+qEHVI achieves 225\% sample efficiency compared to other BO algorithms on Nasbench201. In addition, evolutionary algorithms can be trapped into local optima because they rely on mutation and crossover of previous samples to generate new ones. By using MCTS, LaMOO+CMA-ES can explore the search space between different iterations, greatly improving upon CMA-ES in NasBench201. In short, this result indicates that space partitioning, the core of \ours, leads to faster and better optimization in NAS-based problems.

\subsubsection{Evaluations using NasBench301}

\begin{figure}[t]
\centering 
\subfloat[][Bayesian optimization]{\includegraphics[width=.48\textwidth]{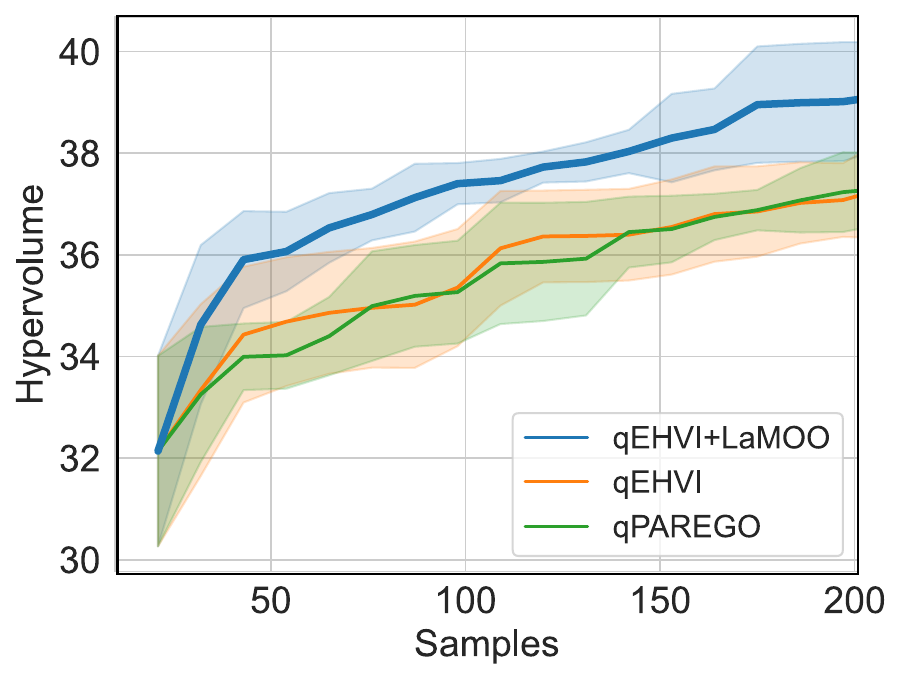}\label{301_baye}} \quad
\subfloat[][Evolutionary search]{\includegraphics[width=.483\textwidth]{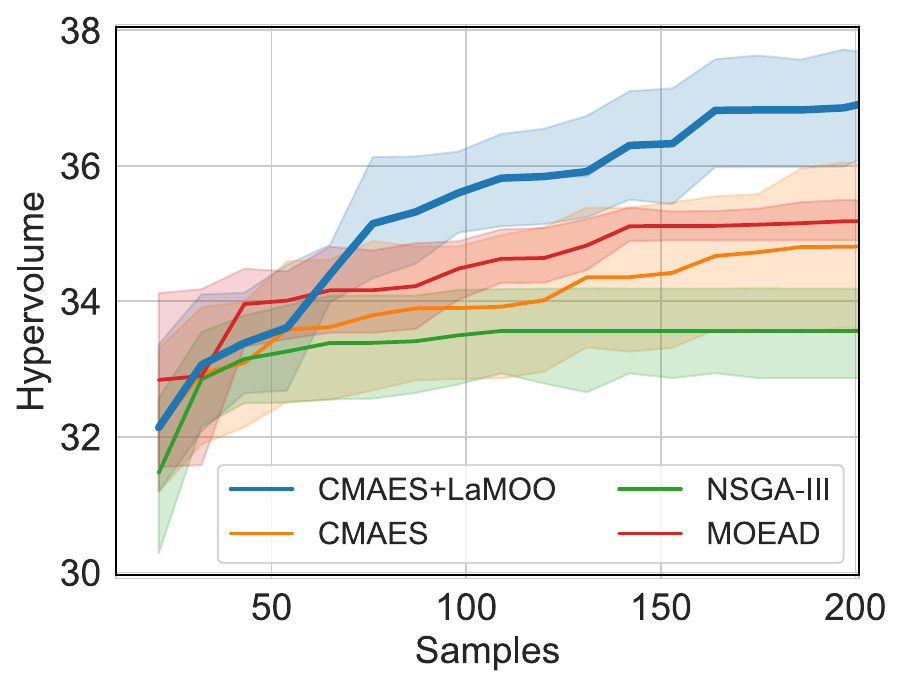}\label{301_evo}}  \quad \\
\caption{The search performance of \ours on NasBench301. We ran each algorithm seven times (the shaded areas are $\pm$ std of the mean).}
\label{fig:nasbench301}
\end{figure}

\begin{figure}[t]
\centering 
\subfloat[][Bayesian optimization]{\includegraphics[width=.475\textwidth]{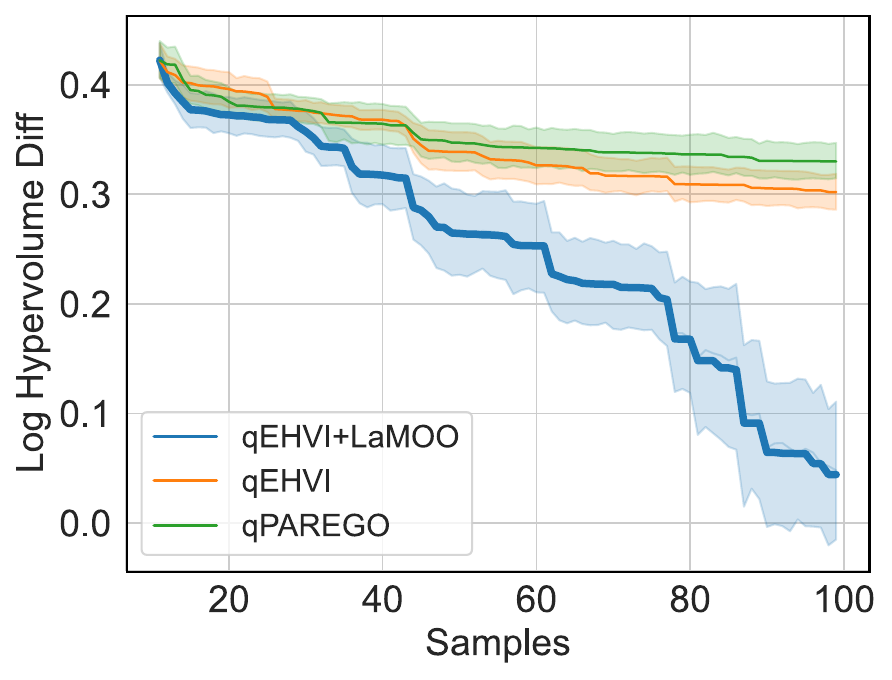}\label{fig:hwnasbench_a}} \quad
\subfloat[][Evolutionary search]{\includegraphics[width=.49\textwidth]{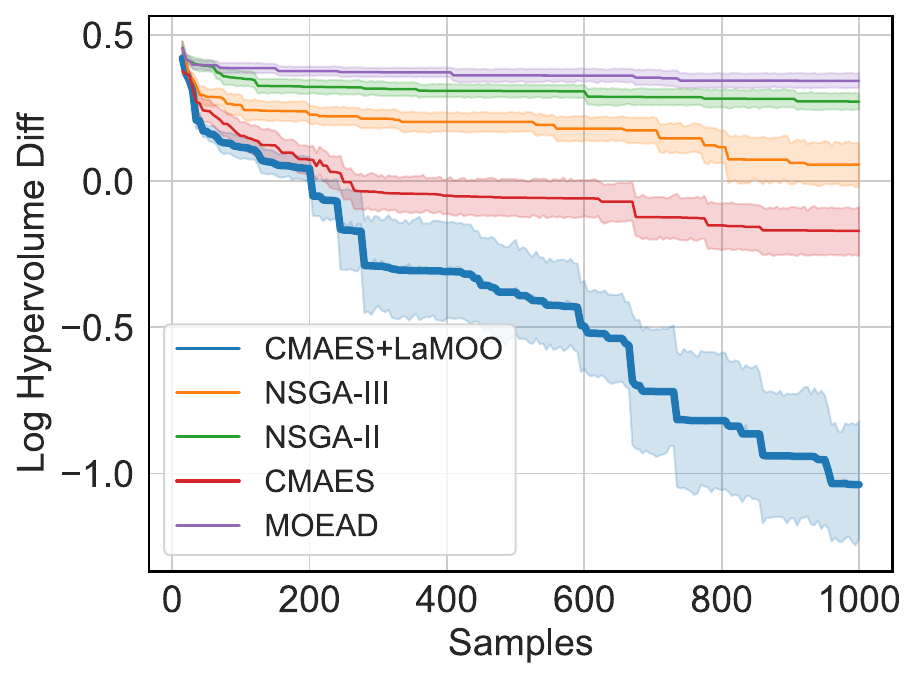}\label{fig:hwnasbench_b}}  \quad \\
\caption{
The search performance of \ours on HW-NAS-BENCH. We ran each algorithm seven times (the shaded areas are $\pm$ std of the mean).
}
\label{fig:hwnasbench}
\end{figure}

\para{Dataset overview.}
NasBench301~\cite{nasbench301} is a surrogate benchmark for the NASNet~\cite{nasnet} search space, which contains more than $10^{21}$ architectures. Specifically, NasBench301 leverages a surrogate model to fit on a subset of architectures ($\sim$60k architectures), and predict the architecture's performance in the entire DARTS search space on the CIFAR10 image classification task. The surrogate model in NasBench301 demonstrates accurate regression results for evaluating the architectures in the large search space. NasBench301 provides the architecture accuracy from the surrogate model. Other basic metrics of architectures, such as \#Params, \#FLOPs, and inference time, can be easily measured in the evaluation process. For this multi-objective optimization, we maximize the inference accuracy and minimize \#Params at the same time in the NasBench301 search space. For NASNet search space, it contains the operations of 3x3 max pool, 3x3, 5x5, depth-separable conv, and skip connection. The search target is to get the architectures for a reduction and a normal cell, and the number of nodes within a cell is 4. This formulates a search space of 3.5 × $10^{21}$ architectures~\cite{nasbench301, lanas}. We use the same encoding method as the NASNet search space for the open-domain CIFAR10 problem and encode the architectures in the NASNet search space with vectors of 16 numbers. Specifically, the first 4 elements represent the operations in a normal cell, 5-8 elements are the concatenation type of the normal cell, 9-12 elements represent the operations in the reduction cell, and the last four elements are the concatenation type of reduction cell. Similar to NasBench201, we normalized \#Params to range $[-1, 0]$ and accuracy to $[0, 1]$.  

\para{Metric.}
Since NasBench301 is a large-scale search space, the maximum hypervolume of this search space is unknown. Instead of using the hypervolume difference, we directly leverage the hypervolume value to demonstrate the performance of the search.

\para{Results.}
As shown in Figure~\ref{fig:nasbench301}, \oursnas with qEHVI outperforms all our BO baselines, and \oursnas with CMA-ES outperforms all our EA baselines, in terms of hypervolume. 
In terms of Bayesian optimization-based algorithms, qEHVI+LaMOO has 200\% search performance improvement compared to qEHVI and qPAREGO. 
For evolutionary-based algorithms, CMAES with \oursnas improves more than 250\% sample efficiency compared to other baselines. It also largely outperforms other baselines after 200 search samples. This result indicates that \oursnas can effectively solve the high-dimensional NAS problems, 16 in the case of Nasbench301.

\subsubsection{Many-Objective Search on Hardware-aware NasBench}

\para{Dataset overview.}
HW-NAS-Bench~\cite{hwnasbench} extends the original NasBench201 dataset by providing additional information that facilitates the consideration of a broader spectrum of metrics (e.g., inference latency, parameter size, \#FLOPs, etc.).
This dataset enables us to evaluate the performance of multi-objective neural architecture search. We adopted the same architecture encoding strategy as NasBench201, as detailed in \S\ref{subsec:nasbench201}.

In our search process, we search for four distinct metrics: accuracy, number of parameters, \#FLOPs, and inference latency on edge GPU. We selected these metrics because they are crucial indicators that model designers tend to prioritize~\cite{nasnet, rea, resnet, gpunet, regnet}. Note that these metrics vary significantly in their range scales due to differing units of measurement. For instance, the highest number of FLOPs recorded is 220 million, whereas the longest inference latency is merely 0.024 seconds. Such disparities in scale could disproportionately bias the search toward metrics with larger numerical values. To mitigate this issue, we employ a min-max normalization technique that re-scales the values to a uniform range between 0 and 1, which allows us to leverage the complete architecture information available within the entire search space.

\para{Metric.}
Similar to NasBench201, we employ the log hypervolume difference to assess the search performance of various algorithms. The specifics of this metric are detailed in \S\ref{subsec:nasbench201}.

\para{Results.}
Figure~\ref{fig:hwnasbench} illustrates a comparison between \oursnas and other baselines. As depicted in Figure~\ref{fig:hwnasbench_a}, our \oursnas, when coupled with qEHVI, exhibits superior performance, significantly surpassing other Bayesian optimization baselines in terms of hypervolume. Figure~\ref{fig:hwnasbench_b} indicates that evolutionary-based search algorithms tend to reach a performance plateau when addressing many-objective NAS problems. However, when \oursnas is paired with CMA-ES, \oursnas helps it escape the initial region to focus on a smaller promising region by space partitioning. Figure~\ref{fig:hwnasbench_b} shows that CMA-ES+\oursnas achieves a search performance increase of over 250\% compared to other evolutionary-based baselines. For example, CMA-ES+\oursnas achieves better hypervolume value with 400 samples than CMA-ES with 1000 samples. These results demonstrate the efficacy of \oursnas in many-objective neural architecture search tasks. 

While \oursnas exhibits strong performance in many-objective NAS tasks, there is a potential limitation in the search process. As the number of objectives increases, even with the implementation of \textit{leaf selection}, computing the hypervolume value for samples within each leaf node of the Monte Carlo Search Tree is challenging. As detailed in \S\ref{subsec:search_promising_region_selection}, the computational cost for hypervolume evaluation grows exponentially. This increase may slow down the search speed, particularly as the number of samples expands with the search progress.

\subsection{Open-Domain NAS Tasks}

To further demonstrate the search performance of \ours on open-domain NAS problems, we use \ours to search for architectures on CIFAR-10 using the NASNet search space, and on ImageNet using the EfficientNet search space. We also leverage \ours to search for architectures on object detection tasks and on Penn Treebank Language models.

\subsubsection{Searching on CIFAR10 Image Classification Task}

\para{Search space overview.}
Our search space aligns with the NASNet search space~\cite{nasnet}. Specifically, it includes a total of eight searchable operations: 3x3 max pooling, 3x3 average pooling, 3x3, 5x5, and 7x7 depthwise convolutions, 3x3 and 5x5 dilated convolutions, and skip connection. The architecture is comprised of normal cells, which maintain the feature map's size, and reduction cells that both upscale the channels by a factor of two and down-sample the resolution of the feature map by two. Each cell incorporates 4 nodes connected by 8 different operations. There are a total of $10^{21}$ architectures in the search space. We use the same encoding strategy as previous work~\cite{lanas}. Each sampled network is trained for 600 epochs, with a batch size of 128, using a momentum SGD optimizer initiated with a learning rate of 0.025, which is then subject to a cosine learning rate schedule throughout the training period. Weight decay is employed for regularization purpose.

\begin{table}[t]
\centering
 \makeatletter\def\@captype{table}\makeatother\caption{Results on CIFAR-10 using the NasNet search space. The two optimization objectives we are searching for are \#params and accuracy. Here, LaMOONet-P1 and LaMOONet-P2 represent architectures searched by LaMOO with smaller and larger parameter sizes, and one-shot LaMOONet and few-shot LaMOONet represent architectures discovered by LaMOO using one-shot and few-shot method. Baseline methods are sorted in descending order of test error. }
\label{tab:cifar}
\resizebox{\textwidth}{!}{%
\begin{tabular}{@{}lllllll@{}}
\toprule
\textbf{Method}                                      & \textbf{Evaluation Method}             & \textbf{\#Parameters}  &\textbf{\#Samples}               & \textbf{Test Error(\%)}                                & \textbf{GPU days}         \\ \midrule
\multicolumn{1}{|l|}{PNASNet-5~\cite{pnas}}                    & \multicolumn{1}{l|}{Standard}  & \multicolumn{1}{r|}{3.2M}  & \multicolumn{1}{r|}{1160}  & \multicolumn{1}{r|}{3.41$\pm$0.09}          & \multicolumn{1}{r|}{225} \\
\multicolumn{1}{|l|}{NAO~\cite{nao}}                    & \multicolumn{1}{l|}{Standard}  & \multicolumn{1}{r|}{3.2M}  &\multicolumn{1}{r|}{1000}   & \multicolumn{1}{r|}{3.14$\pm$0.09}          & \multicolumn{1}{r|}{225} \\
\multicolumn{1}{|l|}{NASNet-A~\cite{nasnet}}                       & \multicolumn{1}{l|}{Standard}   & \multicolumn{1}{r|}{3.3M} & \multicolumn{1}{r|}{20000}  & \multicolumn{1}{r|}{2.65}                   & \multicolumn{1}{r|}{2000} \\
\multicolumn{1}{|l|}{LEMONADE~\cite{lamonade}}                    & \multicolumn{1}{l|}{Standard}  & \multicolumn{1}{r|}{13.1M}  &\multicolumn{1}{r|}{-}   & \multicolumn{1}{r|}{2.58}          & \multicolumn{1}{r|}{90} \\
\multicolumn{1}{|l|}{AlphaX~\cite{alphax}}                     & \multicolumn{1}{l|}{Standard} & \multicolumn{1}{r|}{2.83M}  & \multicolumn{1}{r|}{1000}  & \multicolumn{1}{r|}{2.54$\pm$0.06}                   & \multicolumn{1}{r|}{1000}  \\
\multicolumn{1}{|l|}{AmoebaNet-B-small~\cite{rea}}                    & \multicolumn{1}{l|}{Standard} & \multicolumn{1}{r|}{2.8M} & \multicolumn{1}{r|}{27000}  & \multicolumn{1}{r|}{2.50$\pm$0.05}          & \multicolumn{1}{r|}{3150} \\

\multicolumn{1}{|l|}{\textbf{LaMOONet-P1}} & \multicolumn{1}{l|}{Standard}  & \multicolumn{1}{r|}{\textbf{1.62M}}  & \multicolumn{1}{r|}{600}  & \multicolumn{1}{r|}{2.64$\pm$0.03} & \multicolumn{1}{r|}{100}  \\
\multicolumn{1}{|l|}{\textbf{LaMOONet-P2}} & \multicolumn{1}{l|}{Standard}  & \multicolumn{1}{r|}{3.25M}  & \multicolumn{1}{r|}{600}  & \multicolumn{1}{r|}{\textbf{2.23}$\pm$0.06} & \multicolumn{1}{r|}{100}  \\

\midrule
\multicolumn{1}{|l|}{BayeNAS~\cite{bayesnas}}                          & \multicolumn{1}{l|}{One-shot}  & \multicolumn{1}{r|}{3.4M}   &\multicolumn{1}{r|}{N/A}  & \multicolumn{1}{r|}{2.81$\pm$0.04}          & \multicolumn{1}{r|}{0.2}    \\
\multicolumn{1}{|l|}{DARTS~\cite{DARTS}}                          & \multicolumn{1}{l|}{One-shot}  & \multicolumn{1}{r|}{3.3M}  &\multicolumn{1}{r|}{N/A}   & \multicolumn{1}{r|}{2.76$\pm$0.09}          & \multicolumn{1}{r|}{1.0}    \\
\multicolumn{1}{|l|}{MergeNAS~\cite{mergenas}}                          & \multicolumn{1}{l|}{One-shot}  & \multicolumn{1}{r|}{2.9M}   &\multicolumn{1}{r|}{N/A}  & \multicolumn{1}{r|}{2.68$\pm$0.01}          & \multicolumn{1}{r|}{0.6}    \\
\multicolumn{1}{|l|}{One-shot REA}                  & \multicolumn{1}{l|}{One-shot}  & \multicolumn{1}{r|}{3.5M}  & \multicolumn{1}{r|}{N/A}  & \multicolumn{1}{r|}{2.68$\pm$0.03}          & \multicolumn{1}{r|}{0.75} \\
\multicolumn{1}{|l|}{CNAS~\cite{cnas}}                          & \multicolumn{1}{l|}{One-shot}  & \multicolumn{1}{r|}{3.7M}  & \multicolumn{1}{r|}{N/A} & \multicolumn{1}{r|}{2.60$\pm$0.06}          & \multicolumn{1}{r|}{0.3}    \\
\multicolumn{1}{|l|}{PC-DARTS~\cite{pcdarts}}                          & \multicolumn{1}{l|}{One-shot}  & \multicolumn{1}{r|}{3.6M}   &\multicolumn{1}{r|}{N/A}  & \multicolumn{1}{r|}{2.57$\pm$0.07}          & \multicolumn{1}{r|}{0.3}    \\
\multicolumn{1}{|l|}{Fair-DARTS~\cite{fairdarts}}                          & \multicolumn{1}{l|}{One-shot}  & \multicolumn{1}{r|}{3.32M}   &\multicolumn{1}{r|}{N/A}  & \multicolumn{1}{r|}{2.54$\pm$0.05}          & \multicolumn{1}{r|}{3.0}    \\
\multicolumn{1}{|l|}{ASNG-NAS~\cite{asng-nas}}                          & \multicolumn{1}{l|}{One-shot}  & \multicolumn{1}{r|}{3.32M}   &\multicolumn{1}{r|}{N/A}  & \multicolumn{1}{r|}{2.54$\pm$0.05}          & \multicolumn{1}{r|}{0.11}    \\
\multicolumn{1}{|l|}{P-DARTS~\cite{pdarts}}                          & \multicolumn{1}{l|}{One-shot}  & \multicolumn{1}{r|}{3.4M}   &\multicolumn{1}{r|}{N/A}  & \multicolumn{1}{r|}{2.50}          & \multicolumn{1}{r|}{0.3}    \\

\multicolumn{1}{|l|}{One-shot LaNas~\cite{lanas}}    & \multicolumn{1}{l|}{One-shot}            & \multicolumn{1}{r|}{3.6M}   &\multicolumn{1}{r|}{N/A}  & \multicolumn{1}{r|}{2.24$\pm$0.02}                   & \multicolumn{1}{r|}{3.0}  \\

\multicolumn{1}{|l|}{\textbf{One-shot LaMOONet}} & \multicolumn{1}{l|}{One-shot}  & \multicolumn{1}{r|}{\textbf{1.68M}}  & \multicolumn{1}{r|}{N/A}  & \multicolumn{1}{r|}{2.85$\pm$0.08} & \multicolumn{1}{r|}{1.18}  \\
\multicolumn{1}{|l|}{\textbf{Few-shot LaMOONet}} & \multicolumn{1}{l|}{One-shot}  & \multicolumn{1}{r|}{\textbf{1.65M}}  & \multicolumn{1}{r|}{N/A}  & \multicolumn{1}{r|}{2.78$\pm$0.05} & \multicolumn{1}{r|}{2.06}  \\

\bottomrule
\end{tabular}%

}
\end{table}

\para{Results.}
We apply our \ours with three mainstream NAS evaluation methods, i.e., vanilla NAS, one-shot NAS~\cite{uniform_random_train}, and few-shot NAS~\cite{few-shot}. Table~\ref{tab:cifar} summarizes the SoTA results with DARTS and NASNet search space on CIFAR10, where the first group is models found with vanilla NAS and the second group with one-shot NAS. For \ours, we pick the best architectures from the Pareto frontier, where the accuracies are acquired by actual training for vanilla NAS and supernet(s) estimation for one-shot NAS. 
\ours with qEHVI finds architectures with similar accuracy to ones by vanilla NAS search, with fewer \#Params at 1.62M, compared to at least 2.8M. In addition, in terms of the search cost measured in GPU days, \oursnas takes 30X fewer samples to find the LaMOONet-P2 architecture which achieve better accuracy (97.77\% test accuracy) than AmoebaNet~\cite{rea}, the best performing SoTA architecture found with vanilla NAS. The one-shot and few-shot LaMOONet also demonstrate strong results in terms of both \#Params and accuracy compared to their counterparts.   
The performance gap between the one-shot (second group) and vanilla NAS (first group) methods is because of supernet's poor accuracy prediction~\cite{uniform_random_train, yu2019evaluating}. Few-shot NAS~\cite{few-shot} narrows this performance gap by using multiple supernets, with a slightly increased search cost from 1.18 to 2.06 GPU days.

\subsubsection{Searching on ImageNet Image Classification Task}

The ImageNet search space comes from EffcientNet~\cite{efficientnet}. The depth of an Inverted Residual Block (IRB) can be 2, 3, or 4, along with 3 types of connection patterns; and the expansion ratio within an IRB can be 3, 5, 6, or 7. The kernel size is chosen from {3, 5, 7}. Therefore, the total possible architectures are $((3 * 3 * 4)^{2} + (3 * 3 * 4)^3 + (3 * 3 * 4)^4)^5 \approx 10^{31}$. The details of the connection search space can be found in Appendix~\ref{sec:redesign_irb}. 
For this task, prior work either developed a model on ImageNet that prioritizes accuracy and the number of \#FLOPs~\cite{efficientnet, mobilenetv1, few-shot, lanas, alphax, nasnet} or focused on accuracy and inference latency~\cite{gpunet, ofa}. Consequently, we have conducted two NAS experiments on ImageNet: the first targets accuracy and the number of \#FLOPs, and the second aims for accuracy while considering TensorRT latency with FP16 on an NVIDIA GV100. 
For the TensorRT latency setup, we fixed the TensorRT workspace at 10GB for all runs and benchmarked the latency using a batch size of 1 with explicit shape configuration. We report the average latency derived from 1000 runs. 
Recall that we have demonstrated that one-shot LaMOONet and few-shot LaMOONet achieve similar performance in Table~\ref{tab:cifar}; therefore, we will focus on evaluating the search performance of \oursnas with the one-shot pipeline.

\emph{\textbf{ImageNet training setup.}} For each architecture in the Pareto frontier, we train it using 8 Tesla V100 GPUs with images of a 224x224 resolution in (accuracy, \#FLOPs) two-objective search. For the (accuracy, latency) two-objective search, we set the image resolution of searched architectures as 320x320. We use the standard SGD optimizer with Nesterov momentum 0.9 and set the weight decay to be $3 \times 10^{-5}$. 
Each architecture is trained for a total of 450 epochs, with the first 10 epochs as the warm-up period. During the warm-up epochs, we use a constant learning rate of 0.01. The remaining epochs are trained with an initial learning rate of 0.1, a cosine learning rate decay schedule~\cite{cosine_decay}, and a batch size of 1024 (i.e., 128 images per GPU). The model parameters are subject to a decay factor of 0.9997 to further improve the training performance of our models.

\begin{table}[t]
\centering
 \makeatletter\def\@captype{table}\makeatother\caption{Results on ImageNet using the EfficientNet search space. 
 The two optimization objectives we are searching for are \#FLOPs and accuracy.
 Here, LaMOONet-F1 and LaMOONet-F2 represent architectures searched by LaMOO with smaller and larger \#FLOPs. 
 We report the metric values where applicable based on reported values from original publications.
 Baseline methods are sorted in ascending order of top-1 accuracy. 
 }
 
\label{tab:lamoo_imagenet}
\resizebox{0.82\textwidth}{!}{%
\begin{tabular}{@{}lllll@{}}
\toprule
\textbf{Method}                                      & \textbf{\#FLOPs}              & \textbf{\#Params}          & \textbf{Top-1 Acc(\%)}                                & \textbf{GPU days}         \\ \midrule

\multicolumn{1}{|l|}{REGNETY-400MF~\cite{space_design_1}}                          & \multicolumn{1}{r|}{400M} & \multicolumn{1}{r|}{4.3M}  & \multicolumn{1}{r|}{74.1}          & \multicolumn{1}{r|}{-}    \\
\multicolumn{1}{|l|}{AutoSlim~\cite{autoslim}}                       & \multicolumn{1}{r|}{305M} & \multicolumn{1}{r|}{5.7M}  & \multicolumn{1}{r|}{74.2}                   & \multicolumn{1}{r|}{-} \\
\multicolumn{1}{|l|}{MnasNet-A1~\cite{mnasnet}}                          & \multicolumn{1}{r|}{312M} & \multicolumn{1}{r|}{3.9M}  & \multicolumn{1}{r|}{75.2}          & \multicolumn{1}{r|}{-}    \\
\multicolumn{1}{|l|}{MobileNet-V3-large~\cite{mobilenetv3}}                    & \multicolumn{1}{r|}{219M} & \multicolumn{1}{r|}{5.8M}  & \multicolumn{1}{r|}{75.2}          & \multicolumn{1}{r|}{-} \\
\multicolumn{1}{|l|}{FairDARTS~\cite{fairdarts}}                       & \multicolumn{1}{r|}{440M} & \multicolumn{1}{r|}{4.3M}  & \multicolumn{1}{r|}{75.6}                   & \multicolumn{1}{r|}{3.0} \\
\multicolumn{1}{|l|}{FBNetV2-F4~\cite{fbnetv2}}                    & \multicolumn{1}{r|}{238M} & \multicolumn{1}{r|}{5.6M}  & \multicolumn{1}{r|}{76.0}          & \multicolumn{1}{r|}{8.3} \\
\multicolumn{1}{|l|}{BigNAS~\cite{bignas}}                          & \multicolumn{1}{r|}{242M} & \multicolumn{1}{r|}{4.5M}  & \multicolumn{1}{r|}{76.5}          & \multicolumn{1}{r|}{-}    \\
\multicolumn{1}{|l|}{OFA-small~\cite{alphax}}                     & \multicolumn{1}{r|}{230M} & \multicolumn{1}{r|}{5.4M}  & \multicolumn{1}{r|}{76.9}                   & \multicolumn{1}{r|}{1.6}  \\
\multicolumn{1}{|l|}{MixNet-M~\cite{mixconv}}                          & \multicolumn{1}{r|}{360M} & \multicolumn{1}{r|}{5.0M}  & \multicolumn{1}{r|}{77.0}          & \multicolumn{1}{r|}{-}    \\

\multicolumn{1}{|l|}{EfficientNet-B0~\cite{efficientnet}}                          & \multicolumn{1}{r|}{390M} & \multicolumn{1}{r|}{5.3M}  & \multicolumn{1}{r|}{77.3}          & \multicolumn{1}{r|}{-}    \\
\multicolumn{1}{|l|}{AtomNAS~\cite{atomnas}}                       & \multicolumn{1}{r|}{363M} & \multicolumn{1}{r|}{5.9M}  & \multicolumn{1}{r|}{77.6}                   & \multicolumn{1}{r|}{-} \\
\multicolumn{1}{|l|}{\textbf{LaMOONet-F0}}                          & \multicolumn{1}{r|}{\textbf{248M}} & \multicolumn{1}{r|}{5.1M}  & \multicolumn{1}{r|}{\textbf{78.0}}          & \multicolumn{1}{r|}{1.5}    \\

\midrule
\multicolumn{1}{|l|}{ChamNet~\cite{chamnet}}                          & \multicolumn{1}{r|}{553M} & \multicolumn{1}{r|}{-}  & \multicolumn{1}{r|}{75.4}          & \multicolumn{1}{r|}{-}    \\
\multicolumn{1}{|l|}{RegNet~\cite{regnet}}                          & \multicolumn{1}{r|}{600M} & \multicolumn{1}{r|}{6.1M}  & \multicolumn{1}{r|}{75.5}          & \multicolumn{1}{r|}{-}    \\
\multicolumn{1}{|l|}{REGNETY-800MF~\cite{space_design_1}}                          & \multicolumn{1}{r|}{800M} & \multicolumn{1}{r|}{6.3M}  & \multicolumn{1}{r|}{76.3}          & \multicolumn{1}{r|}{-}    \\

\multicolumn{1}{|l|}{MixNet-L~\cite{mixconv}}                        & \multicolumn{1}{r|}{565M} & \multicolumn{1}{r|}{7.3M}  & \multicolumn{1}{r|}{78.9}          & \multicolumn{1}{r|}{-}    \\

\multicolumn{1}{|l|}{FBNetV3~\cite{fbnetv3}}                          & \multicolumn{1}{r|}{544M} & \multicolumn{1}{r|}{-}  & \multicolumn{1}{r|}{79.5}          & \multicolumn{1}{r|}{-}    \\

\multicolumn{1}{|l|}{OFA-large~\cite{ofa}}                          & \multicolumn{1}{r|}{595M} & \multicolumn{1}{r|}{9.1M}  & \multicolumn{1}{r|}{80.0}          & \multicolumn{1}{r|}{1.6}    \\

\multicolumn{1}{|l|}{EfficientNet-B2~\cite{efficientnet}}                          & \multicolumn{1}{r|}{1000M} & \multicolumn{1}{r|}{6.1M}  & \multicolumn{1}{r|}{80.3}          & \multicolumn{1}{r|}{-}    \\
\multicolumn{1}{|l|}{NSGANetV2-xl~\cite{nsganetv2}}                          & \multicolumn{1}{r|}{593M} & \multicolumn{1}{r|}{8.7M}  & \multicolumn{1}{r|}{80.4}          & \multicolumn{1}{r|}{1}    \\
\multicolumn{1}{|l|}{\textbf{LaMOONet-F1}} & \multicolumn{1}{r|}{\textbf{522M}} & \multicolumn{1}{r|}{7.8M}  & \multicolumn{1}{r|}{\textbf{80.4}} & \multicolumn{1}{r|}{1.5}  \\

\bottomrule
\end{tabular}%

}
\end{table}

\begin{table}[t]
\centering

 \makeatletter\def\@captype{table}\makeatother\caption{ Results on ImageNet using the EfficientNet search space. The two optimization objectives we are searching for are TensorRT Latency with FP16 in NVIDIA GV100 and accuracy. Here, LaMOONet-G0 and LaMOONet-G1 represent architectures searched by LaMOO with smaller and larger inference latency. 
  Baseline methods are sorted in ascending order of top-1 accuracy. 
 }
 
\label{tab:lamoo_imagenet_latency}
\resizebox{0.96\textwidth}{!}{%
\begin{tabular}{@{}lrrrr@{}}
\toprule
\textbf{Method}                             & \multicolumn{1}{l}{\textbf{Top-1 Acc(\%)}} & \multicolumn{1}{l}{\textbf{\begin{tabular}[c]{@{}l@{}}TensorRT Latency \\ FP16 GV100 (ms)\end{tabular}}} & \multicolumn{1}{l}{\textbf{\begin{tabular}[c]{@{}l@{}}LaMOONet\\ Speedup$\uparrow$\end{tabular}}} & \multicolumn{1}{l}{\textbf{\begin{tabular}[c]{@{}l@{}}LaMOONet\\ Acc Improvement$\uparrow$\end{tabular}}} \\ \midrule
\multicolumn{1}{|l|}{RegNet-X~\cite{regnet}}              & \multicolumn{1}{r|}{77.0}                  & \multicolumn{1}{r|}{2.06}                                                                                & \multicolumn{1}{r|}{3.6x}                                                                             & \multicolumn{1}{r|}{2.2}                                                                              \\
\multicolumn{1}{|l|}{EfficientNet-B0~\cite{efficientnet}}       & \multicolumn{1}{r|}{77.1}                  & \multicolumn{1}{r|}{1.18}                                                                                & \multicolumn{1}{r|}{2.1x}                                                                             & \multicolumn{1}{r|}{2.1}                                                                              \\
\multicolumn{1}{|l|}{FBNetV2-L1~\cite{fbnetv2}}            & \multicolumn{1}{r|}{77.2}                  & \multicolumn{1}{r|}{1.13}                                                                                & \multicolumn{1}{r|}{2.0x}                                                                             & \multicolumn{1}{r|}{2.0}                                                                              \\
\multicolumn{1}{|l|}{EfficientNetX-B0-GPU~\cite{efficientnet}}  & \multicolumn{1}{r|}{77.3}                  & \multicolumn{1}{r|}{1.05}                                                                                & \multicolumn{1}{r|}{1.8x}                                                                             & \multicolumn{1}{r|}{1.9}                                                                              \\

\multicolumn{1}{|l|}{GPUNet-0~\cite{gpunet}}              & \multicolumn{1}{r|}{78.9}                  & \multicolumn{1}{r|}{0.62}                                                                                & \multicolumn{1}{r|}{1.1x}                                                                             & \multicolumn{1}{r|}{0.3}                                                                              \\
\multicolumn{1}{|l|}{\textbf{LaMOONet-G0}} & \multicolumn{1}{r|}{\textbf{79.2}}                      & \multicolumn{1}{r|}{\textbf{0.57}}                                                                                    & \multicolumn{1}{r|}{-}                                                                             & \multicolumn{1}{r|}{-}                                                                              \\ \midrule

\multicolumn{1}{|l|}{FBNetV3-B~\cite{fbnetv3}}             & \multicolumn{1}{r|}{79.8}                  & \multicolumn{1}{r|}{1.55}                                                                                & \multicolumn{1}{r|}{2.2x}                                                                             & \multicolumn{1}{r|}{0.8}                                                                              \\

\multicolumn{1}{|l|}{EfficientNetX-B2-GPU~\cite{efficientnet}}  & \multicolumn{1}{r|}{80.0}                  & \multicolumn{1}{r|}{1.61}                                                                                & \multicolumn{1}{r|}{2.2x}                                                                             & \multicolumn{1}{r|}{0.6}                                                                              \\
\multicolumn{1}{|l|}{RegNet-X~\cite{regnet}}              & \multicolumn{1}{r|}{80.0}                  & \multicolumn{1}{r|}{3.9}                                                                                 & \multicolumn{1}{r|}{5.4x}                                                                             & \multicolumn{1}{r|}{0.6}                                                                              \\
\multicolumn{1}{|l|}{EfficientNet-B2~\cite{efficientnet}}       & \multicolumn{1}{r|}{80.3}                  & \multicolumn{1}{r|}{1.86}                                                                                & \multicolumn{1}{r|}{2.6x}                                                                             & \multicolumn{1}{r|}{0.3}                                                                              \\

\multicolumn{1}{|l|}{ResNet-50~\cite{resnet}}             & \multicolumn{1}{r|}{80.3}                  & \multicolumn{1}{r|}{1.1}                                                                                 & \multicolumn{1}{r|}{1.5x}                                                                             & \multicolumn{1}{r|}{0.3}                                                                              \\
\multicolumn{1}{|l|}{GPUNet-1~\cite{gpunet}}              & \multicolumn{1}{r|}{80.5}                  & \multicolumn{1}{r|}{0.82}                                                                                & \multicolumn{1}{r|}{1.1x}                                                                             & \multicolumn{1}{r|}{0.1}                                                                              \\
\multicolumn{1}{|l|}{\textbf{LaMOONet-G1}} & \multicolumn{1}{r|}{\textbf{80.6}}                      & \multicolumn{1}{r|}{\textbf{0.72}}                                                                                    & \multicolumn{1}{r|}{-}                                                                             & \multicolumn{1}{r|}{-}                                                                              \\ \bottomrule
\end{tabular}
}
\end{table}

\para{Results.} Table \ref{tab:lamoo_imagenet} compares our \oursnas with others SoTA baselines with different \#FLOPs scales. The results demonstrate that the searched architectures by \oursnas greatly outperform other baselines in terms of both \#FLOPs and accuracy.
We group state-of-the-art models by their \#FLOPs. 
The models in the first group have \#FLOPs less than 500M while the models in the second group have \#FLOPs more than 500M. We pick our models, labeled LaMOONet-F0 and LaMOONet-F1, from the Pareto frontier based on the two objectives, accuracy and \#FLOPs. We see that LaMOONet-F1 has the highest accuracy, 0.1 higher than EfficientNet-B2, with only 52.2\% \#FLOPs. Similarly, LaMOONet-F0 also has the highest accuracy in its group, with the lowest \#FLOPs. Table \ref{tab:lamoo_imagenet_latency} compares our models found by \oursnas with other SoTA baselines with different inference latencies. 
The results show that the architectures found by \oursnas significantly surpass all baselines, delivering higher accuracy with reduced TensorRT inference time. Our LaMOONet-G0 achieves a 1.1X speedup over the SoTA architecture GPUNet-0~\cite{gpunet} while having a 0.3\% higher accuracy. LaMOONet-G1 has the highest top-1 accuracy of 80.6\% while incurring the lowest TensorRT latency with FP16 on an NVIDIA GV100.

\subsubsection{Searching on PENN TREEBANK}
We evaluate \oursnas on Penn Treebank (PTB), a widely-studied benchmark for language models. We used
the same search space and training setup as the original DARTS to search RNN on PTB. Here we search for two objectives, i.e., perplexity and \#Params. By using \oursnas, we
achieved the state-of-the-art test Perplexity of 54.56 with only 22M \#Params. In comparison, the one-shot version DARTS algorithm found an architecture with worse performance (55.7 test Perplexity) and more \#Params (23M); the few-shot version DARTS~\cite{few-shot} found an architecture with slightly better performance (54.89 test perplexity) but requires 23M \#Params.

\subsubsection{Searching on MS COCO Object Detection Task}

\begin{figure}[t]
\centering 
\subfloat[][The structure of Monte-Carlo Tree at the final search iteration.]{\includegraphics[width=.46\textwidth]{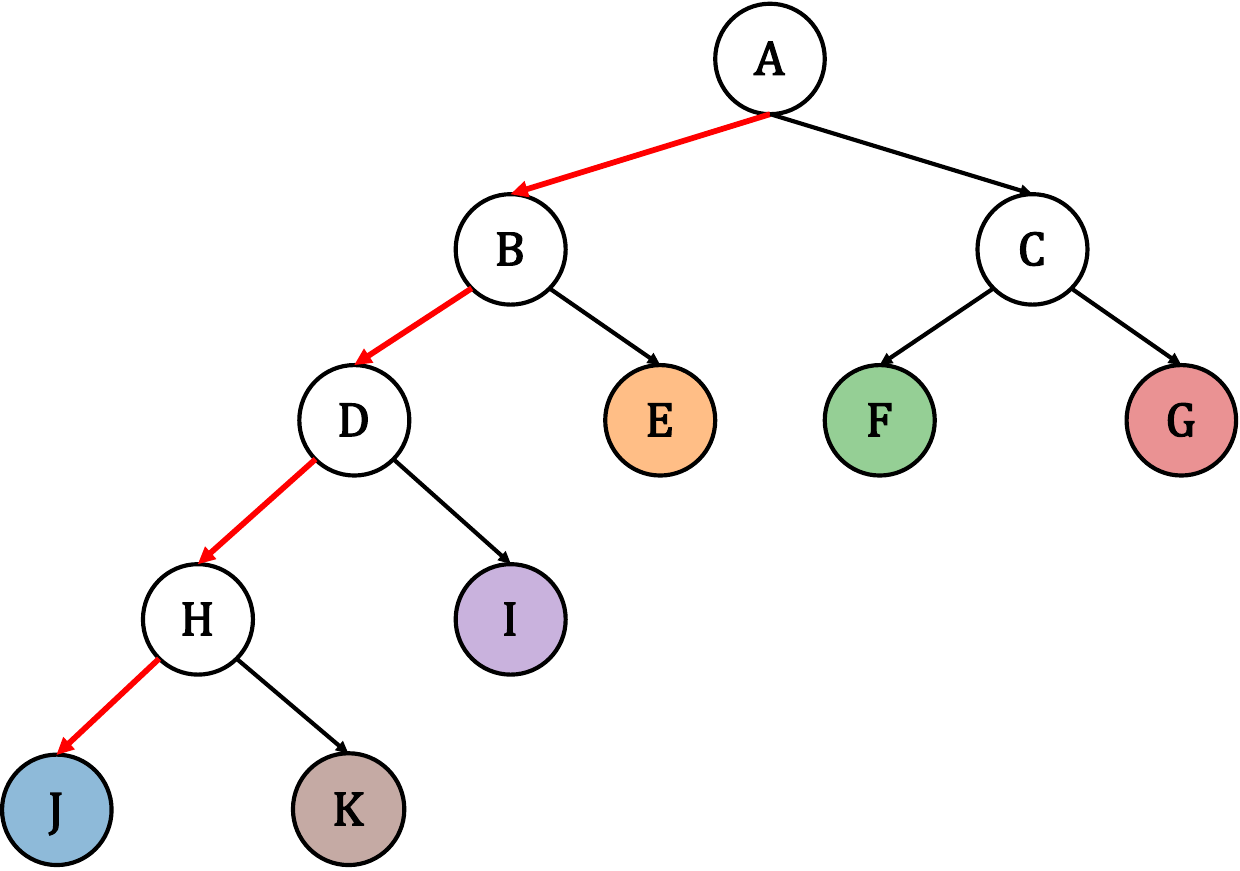}\label{fig:verf_a}} \quad
\subfloat[][The samples and SVM split region of leaf node of the tree in the search space.]{\includegraphics[width=.46\textwidth]{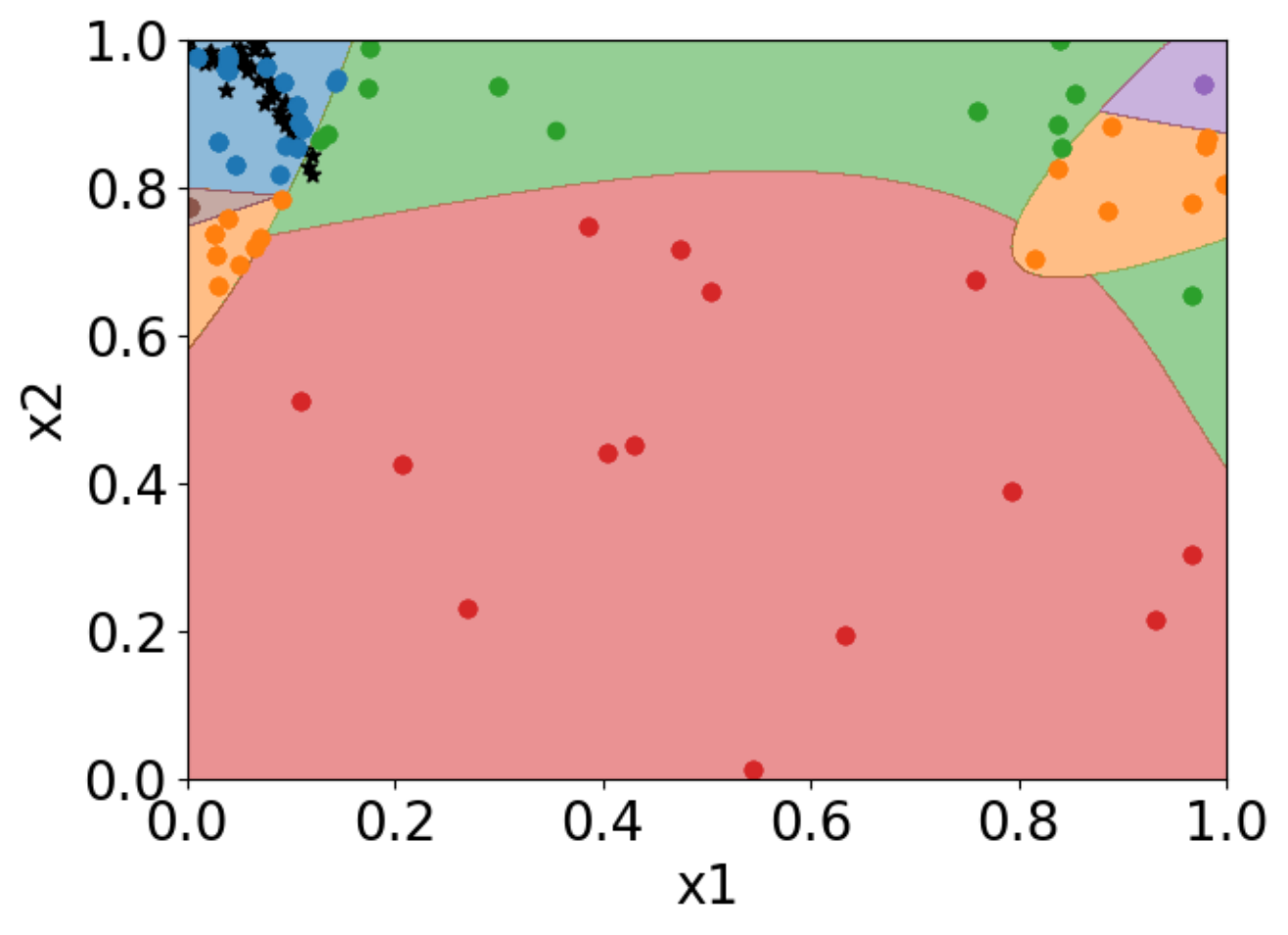}\label{fig:verf_b}}  \quad \\
\subfloat[][The selected region in selected path of the tree.]{\includegraphics[width=.46\textwidth]{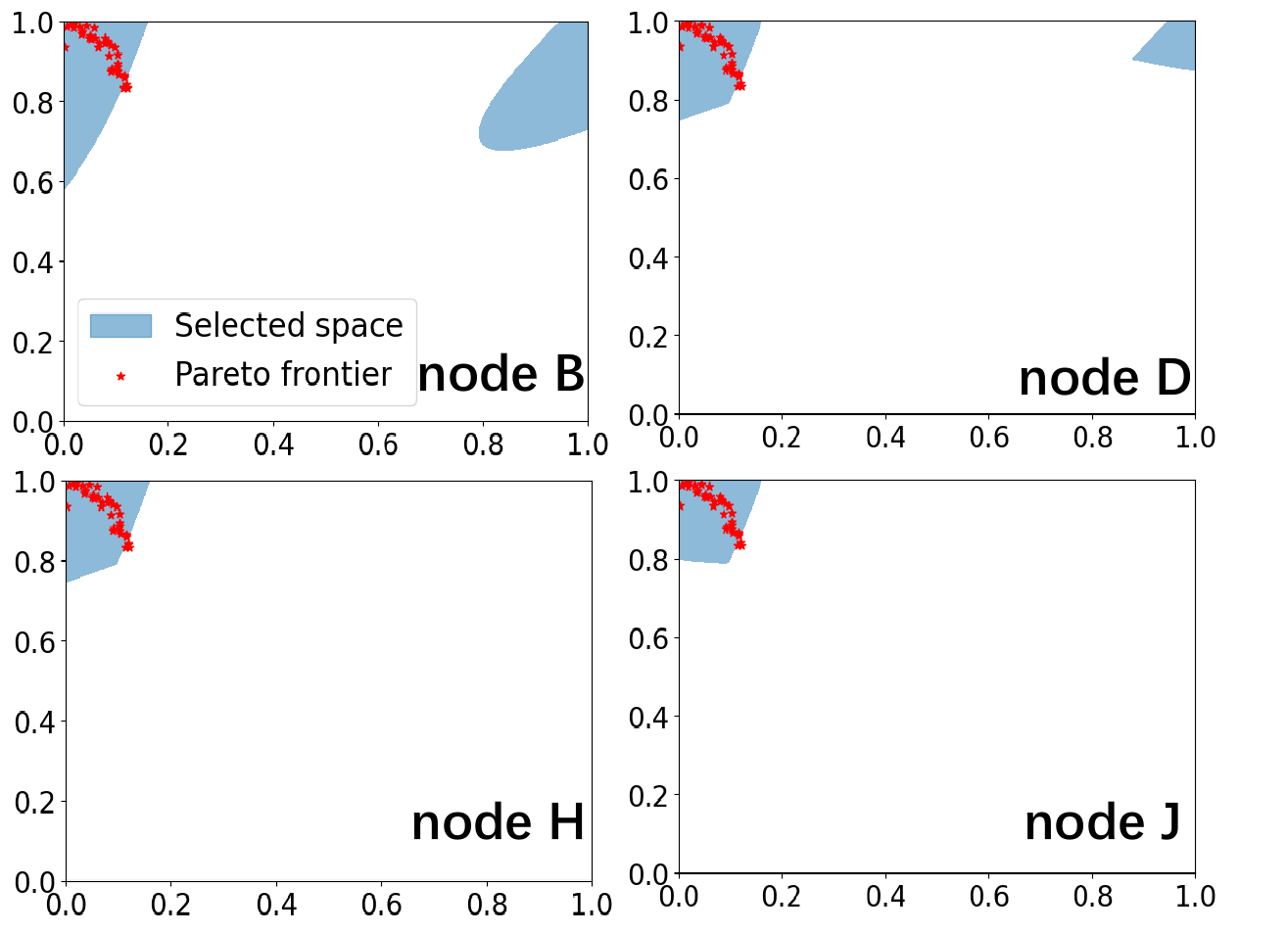}\label{fig:verf_c}} \quad
\subfloat[][The samples in the objective space at different search iterations.]{\includegraphics[width=.46\textwidth]{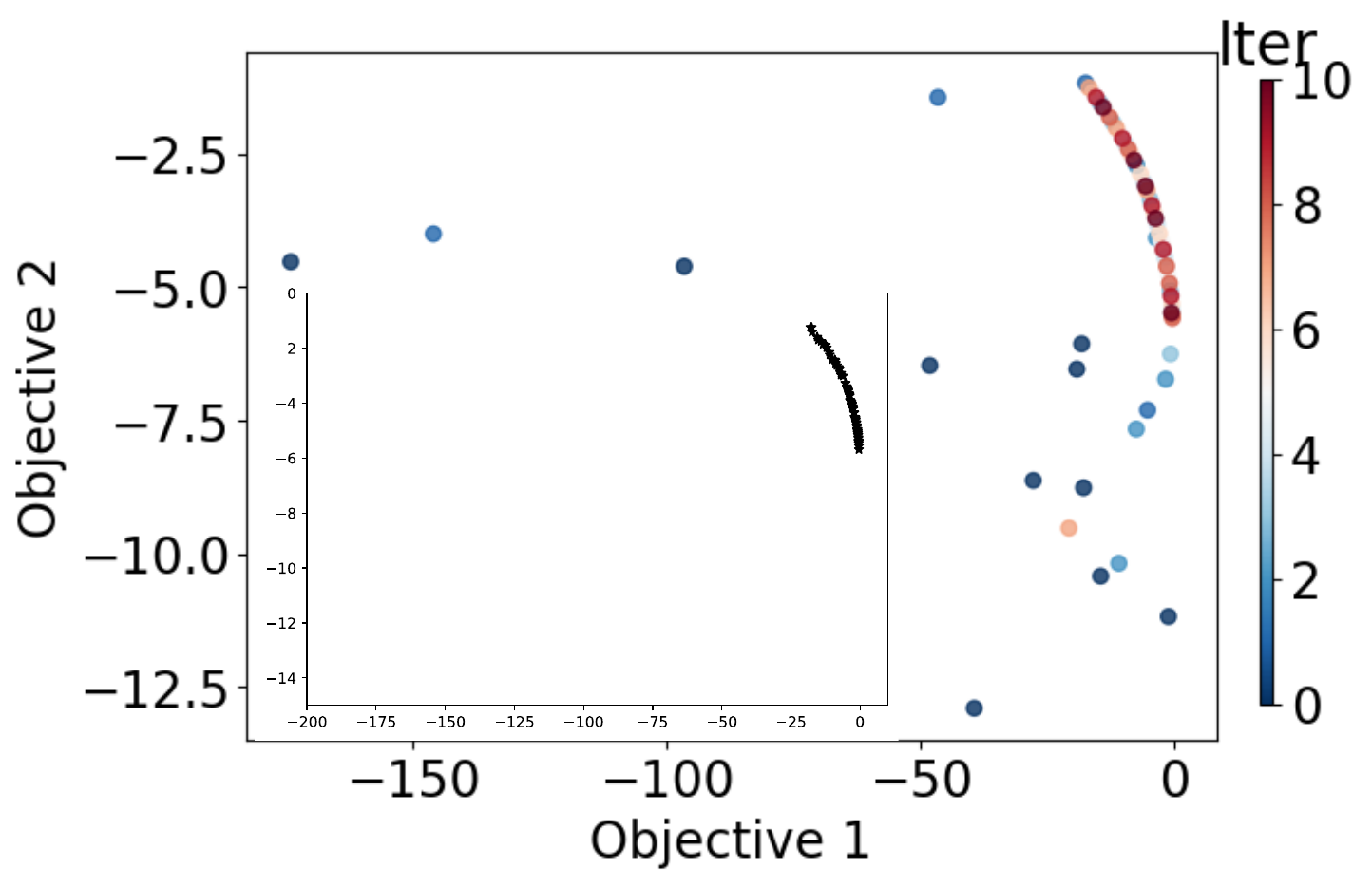}\label{fig:verf_d}}  \quad \\
\caption{
\small Visualization of selected region at different search iterations and nodes. (a) The Monte-Carlo tree with colored leaves. The selected path is marked in red. (b) Visualization of the regions ($\Omega_{J}, \Omega_{K}, \Omega_{I}, \Omega_{E}, \Omega_{F}, \Omega_{G}$) that are consistent with leaves in (a) in the search space. Black dots represent the Pareto frontier estimated by $10^6$ random samples. 
(c) Visualization of the selected path at final iteration.  (d) Visualization of samples during search; bottom left is the Pareto frontier estimated from one million samples.
}
\label{fig:verf}
\end{figure}

\begin{figure}[t]
\centering 
\subfloat[][]{\includegraphics[width=.45\textwidth]{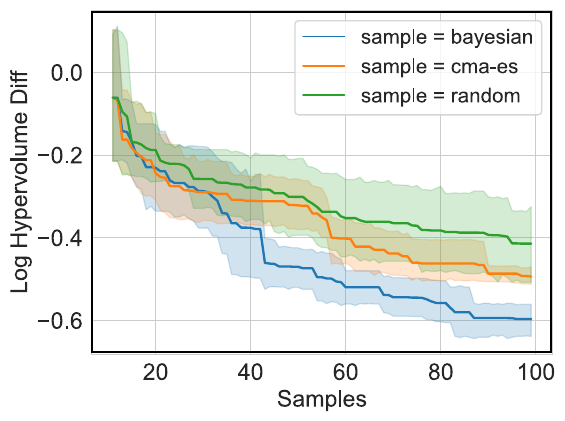}\label{fig:ablation_a}} \quad
\subfloat[][]{\includegraphics[width=.45\textwidth]{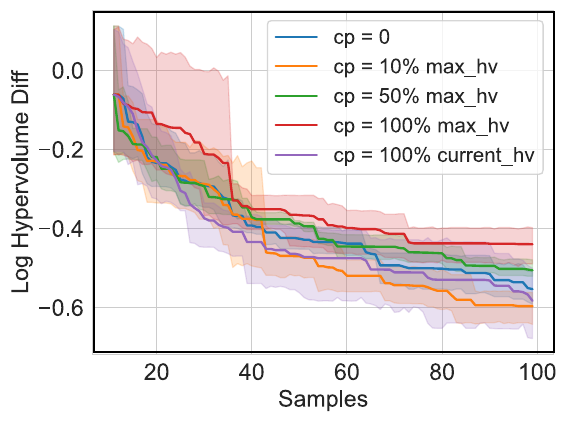}\label{fig:ablation_b}}  \quad \\
\subfloat[][]{\includegraphics[width=.45\textwidth]{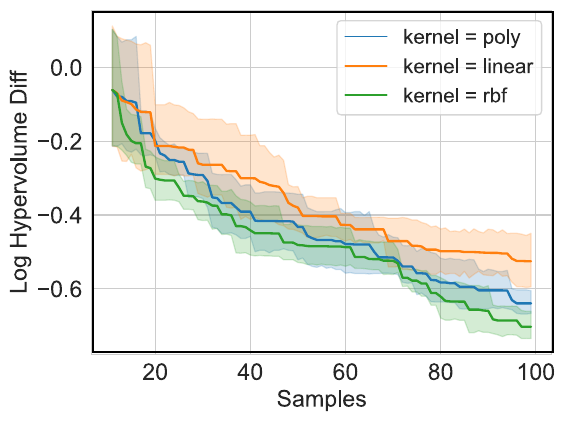}\label{fig:ablation_c}} \quad
\subfloat[][]{\includegraphics[width=.45\textwidth]{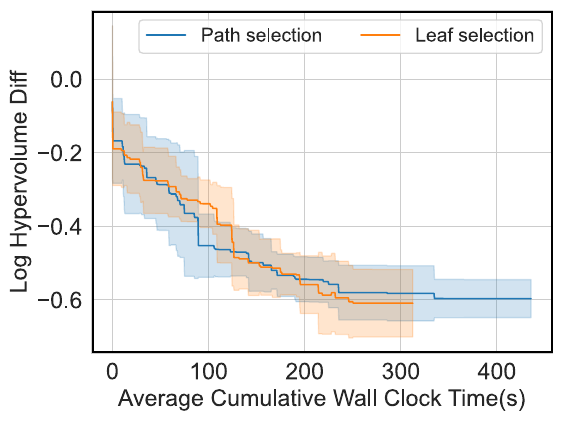}\label{fig:ablation_d}}  \quad \\
\caption{
\small Ablation studies on hyperparameters and sampling methods in \ours on NasBench201. (a) 
Sampling with different methods.
(b) Sampling with different $C_p$. (c) Partitioning with different SVM kernels. (d) Comparison of leaf selection and path selection methods in MCTS.
}
\label{fig:ablation}
\end{figure}

\begin{table}[t]
\caption{Results on the test-dev set of MS COCO with different decoder, backbone, and channels. R-50 represent ResNet50 architecture~\cite{resnet}. Note that NAS-FCOS is the upgraded version of FCOS but we did not implement this on our backbone model. All networks have the same input image resolution.
}
\centering
\resizebox{0.9\textwidth}{!}{
\begin{tabular}{@{}lllll@{}}
\toprule
\textbf{Decoder}                                                     & \textbf{Backbone}                                                         & \textbf{\#Channels}      & \textbf{\#FLOPs(G)}                   & \textbf{AP}                        \\ \midrule
\multicolumn{1}{|l|}{FPN-FCOS~\cite{fcos}}     & \multicolumn{1}{l|}{R-50~\cite{resnet}}             & \multicolumn{1}{r|}{256} & \multicolumn{1}{r|}{169.9}          & \multicolumn{1}{r|}{37.4}          \\ \midrule
\multicolumn{1}{|l|}{NAS-FCOS~\cite{nas-fcos}} & \multicolumn{1}{l|}{MobileNetV2~\cite{mobilenetv2}} & \multicolumn{1}{r|}{128} & \multicolumn{1}{r|}{39.3}           & \multicolumn{1}{r|}{32.0}          \\
\multicolumn{1}{|l|}{NAS-FCOS~\cite{nas-fcos}} & \multicolumn{1}{l|}{MobileNetV2~\cite{mobilenetv2}} & \multicolumn{1}{r|}{256} & \multicolumn{1}{r|}{121.8}          & \multicolumn{1}{r|}{34.7}          \\
\multicolumn{1}{|l|}{FPN-FCOS~\cite{fcos}}     & \multicolumn{1}{l|}{MobileNetV2~\cite{mobilenetv2}} & \multicolumn{1}{r|}{256} & \multicolumn{1}{r|}{105.4}          & \multicolumn{1}{r|}{31.2}          \\ \midrule
\multicolumn{1}{|l|}{FPN-FCOS~\cite{fcos}}     & \multicolumn{1}{l|}{\textbf{LaMOONet-D0}}                  & \multicolumn{1}{r|}{128} & \multicolumn{1}{r|}{\textbf{35.2}}  & \multicolumn{1}{r|}{\textbf{36.5}} \\
\multicolumn{1}{|l|}{FPN-FCOS~\cite{fcos}}     & \multicolumn{1}{l|}{\textbf{LaMOONet-D1}}                  & \multicolumn{1}{r|}{256} & \multicolumn{1}{r|}{\textbf{109.5}} & \multicolumn{1}{r|}{\textbf{37.6}} \\ \bottomrule
\end{tabular}
}

\label{table:lamoo_detection}
\end{table}

We leverage \oursnas searching for the efficient backbone with FPN-FCOS~\cite{fcos} decoder (neck) in object detection. We also compared our searched architectures with the more effective FCOS-based decoder called NAS-FCOS~\cite{nas-fcos}, which leverages NAS for searching in promising FPN-FCOS architectures and has shown better results compared to FPN-FCOS. 
We use our searched efficient backbones and compare the performance to the lightweight backbone MobileNetV2 and a more powerful but non-efficient based backbone ResNet50 (R-50). We implement both 128 and 256 channels of the decoder. The results on the MS COCO test set are shown in Table~\ref{table:lamoo_detection}. 

\emph{\textbf{Object detection training setup.}} We use 4 Telsa V100 GPUs for training our models. We use the standard SGD optimizer with an initial learning rate of 0.005 and a norm gradient clip at 35. We train each model for 24 epochs and use the first 500 iterations as the warm-up phase. During the warm-up iterations, the learning rate starts at 0.002 and increases by 0.00033 every 50 iterations until it reaches 0.005. After the warm-up phase, we divide the learning rate by 10 at the $10^{th}$ and $22^{nd}$ epochs (i.e., to $5 \times 10^{-4}$ and $5 \times 10^{-5}$ respectively). We resize each image to 1333 by 800 and implement a random flip with a 0.5 flip ratio. 
Further, we apply the \emph{center-ness} and \emph{center} sampling techniques on our training pipeline, based on prior work~\cite{fcos, nas-fcos} to further improve the results.

\para{Results.}
Our searched backbone with 256 channels outperforms the ResNet50 with the same channel numbers and decoder by 0.2 AP (average precision) but with 60.4G fewer \#FLOPs. In particular, compared to the mobile-based backbone, which shares the same design style with us, by using FPN-FCOS with 128 channels, LaMOONet-D0 only requires 35.2G \#FLOPs but achieves a promising AP with this scale of \#FLOPs. These results greatly surpass the MobileNetV2 with 256 channels and a better decoder NAS-FCOS. 

In summary, we show that searching with \oursnas leads to architectures of better performance, measured in average precision, with low \#FLOPs. In other words, \oursnas searched models have the potential to run on resource-constrained devices and achieve good detection accuracy for the object detection task.

\subsection{Ablation Studies}

\subsubsection{Visualization of LaMOO}

To gain insights into the operation of \ours, we present a visualization of its optimization process for the Branin-Currin problem, a two-dimensional input problem, for ease of visualization. First, the Pareto optimal set $\Omega_P$ is estimated from $10^6$ random samples (marked as black stars), as shown in both search and objective space (Figure.~\ref{fig:verf_b} and bottom left of Figure~\ref{fig:verf_d}. 
Over several iterations, \ours progressively prunes away unpromising regions so that the remaining regions approach $\Omega_P$.
Figure~\ref{fig:verf_c} shows that the selected nodes consist of promising regions. Figure~\ref{fig:verf_a} shows the final tree structure. The color of each leaf node corresponds to a region in the search space depicted in Figure~\ref{fig:verf_b}. The selected region is recursively bounded by SVM classifiers corresponding to nodes on the selected path (red arrows in Figure~\ref{fig:verf_a}). In the most promising region, $\Omega_{J}$, we could implement any of the NAS and black-box optimization algorithms, such as reinforcement learning, evolutionary algorithms, and Bayesian optimization, to generate new samples. For a given optimization algorithm, we utilize SVM classifiers to define boundaries for the optimization algorithm, therefore confining the generation of new samples within $\Omega_{J}$. This approach effectively enhances sample efficiency by focusing efforts on the most promising areas. \S\ref{subsec:search_sampling_from_a_leaf} provides details of integration \ours with different optimization algorithms.

\subsubsection{Ablation of Design Choices}

We assess the impact of various hyperparameters and sampling methods on the performance of \oursnas. This study is conducted using the NasBench201 dataset as outlined below.  

\emph{\textbf{Sampling methods}}. \oursnas can be combined with different sampling methods, including Bayesian Optimization (such as qEHVI) and evolutionary algorithms (such as CMA-ES). Figure~\ref{fig:ablation_a} indicates that qEHVI substantially enhances performance when compared to random sampling, whereas CMA-ES yields only a slight improvement. These results are in line with our previous discovery that, in the context of MOO, BO generally outperforms EA in terms of search efficiency.

\emph{\textbf{The exploration factor $C_p$}} controls the balance of exploration and exploitation. A larger $C_p$ guides LaMOO to visit the sub-optimal regions more often. Based on the results in Figure~\ref{fig:ablation_b}, greedy search ($C_p=0$) leads to worse performance compared to a proper $C_p$ value (i.e. 10\% of maximum hypervolume), which justifies our usage of MCTS. On the other hand, over-exploration can also yield even worse results than a greedy search. Therefore, a rule of thumb is to set the $C_p$ to be roughly 10\% of the maximum hypervolume $HV_{\max}$. When $HV_{\max}$ is unknown, $C_p$ can be dynamically set to 10\% of the hypervolume of current samples in each search iteration. The final performances by using 10\% HVmax and 10\% current hypervolume are similar.

\emph{\textbf{SVM kernels}}. As shown in Figure~\ref{fig:ablation_c}, we find that the RBF kernel performs the best, in agreement with ~\citep{wang2020learning}. Thanks to the non-linearity of the polynomial and RBF kernels, their region partitions perform better compared to the linear one. 

\emph{\textbf{MCTS node selection}}. We compare the leaf selection and path selection methods in MCTS. Figure~\ref{fig:ablation_d} shows that leaf selection significantly saves search time while achieving similar search performance.

\section{Conclusion and Future Work}

This work applies a novel multi-objective optimizer called \ours to the domain of neural architecture search. We demonstrate that \ours can be seamlessly integrated with three prevalent NAS evaluation methods, i.e., one-shot, few-shot, and performance predictor-based NAS. Through a series of comprehensive experiments, we highlight the superior sample efficiency of \ours compared to existing methodologies for various NAS tasks, including open-source NAS datasets and open-domain NAS tasks. For example, \ours has found state-of-the-art models on ImageNet with an impressive top-1 accuracy of 80.4\% at 522M \#FLOPS and 78.0\% top-1 accuracy at 248 M \#FLOPS. Additionally, on CIFAR10, \ours successfully searched an architecture that delivers 97.36\% top-1 accuracy with only 1.62M \#parameters. These compelling results collectively highlight the efficacy and potential of \ours as a significant tool in multi-objective neural architecture search.

Despite the great success we have demonstrated with \ours in NAS, there are still many fruitful directions to pursue. For example, in this work, we utilize an SVM classifier to partition the search space. Beyond SVM, numerous other classification models, such as deep neural networks, can be integrated into \ours to potentially further enhance the quality of space partitioning. Moreover, as a multi-objective meta-optimizer, \ours can be extended to various research domains beyond NAS, including molecule discovery~\cite{lamoo}, hyperparameter tuning~\cite{ottertune, restune}, optimization of large language models~\cite{llm_personalized}, and other real-world applications.

\section{Acknowledgments}

We thank the anonymous reviewers for their constructive reviews. This work was supported in part by NSF Grants \#2105564 and \#2236987, a VMware grant, the Worcester Polytechnic Institute’s Computer Science Department. Most results presented in this paper were obtained using CloudBank~\cite{cloudBank}, which is supported by the National Science Foundation under award \#1925001. This work also used Expanse at San Diego Supercomputer Center through allocation CIS230364 from the Advanced Cyberinfrastructure Coordination Ecosystem: Services \& Support (ACCESS) program, which is supported by National Science Foundation grants \#2138259, \#2138286, \#2138307, \#2137603, and \#2138296.

\newpage

\appendix
\section{Redesign Connection Pattern in Inverted Residual Block for EfficientNet Search Space}
\label{sec:redesign_irb}

\subsection{Connection search space in Inverted Residual Block}

\begin{figure}[H] 
\centering 
\includegraphics[width=0.8\columnwidth]{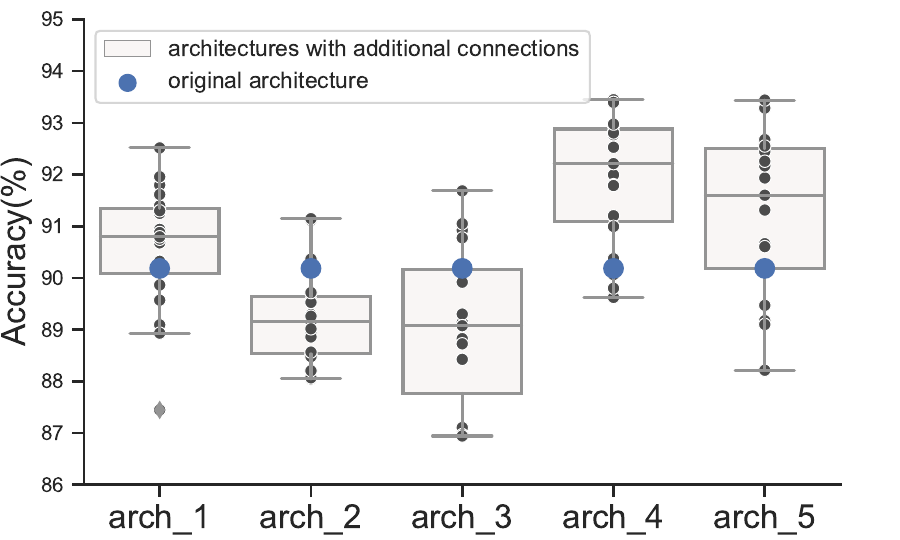}
\caption{Accuracy analysis of Nasbench101~\cite{nasbench101} architectures with different connection patterns. We add different connections into the five sampled architectures (blue circle) and plot the corresponding accuracy distributions. }
\label{fig:compare}
\end{figure}

Existing work on efficient neural networks usually focuses on designing or modifying operation types of the IRB but often overlook another important design factor---the connection pattern~\cite{mobilenetv2, mobilenetv3, fbnet, fbnetv2, ofa, proxylessnas}. As demonstrated by prior works~\cite{DARTS, alphax, nasbench101, nasbench201, nasnet} and also observed by our own analysis (see Figure~\ref{fig:compare}), the connection pattern of a neural architecture plays a crucial role in determining the model accuracy. 

We randomly sample five architectures from the Nasbench101 dataset~\cite{nasbench101}, and use them as the basis for analyzing the impact of connections on model accuracy. Specifically, we add connections to the sampled architectures and plot the corresponding accuracy in Figure~\ref{fig:compare}. We observe a wide range of accuracy distribution (e.g., can be up to tens percent difference) when architectures only differ in connections.

\begin{figure}[H]
\centering 
\includegraphics[width=0.7\columnwidth]{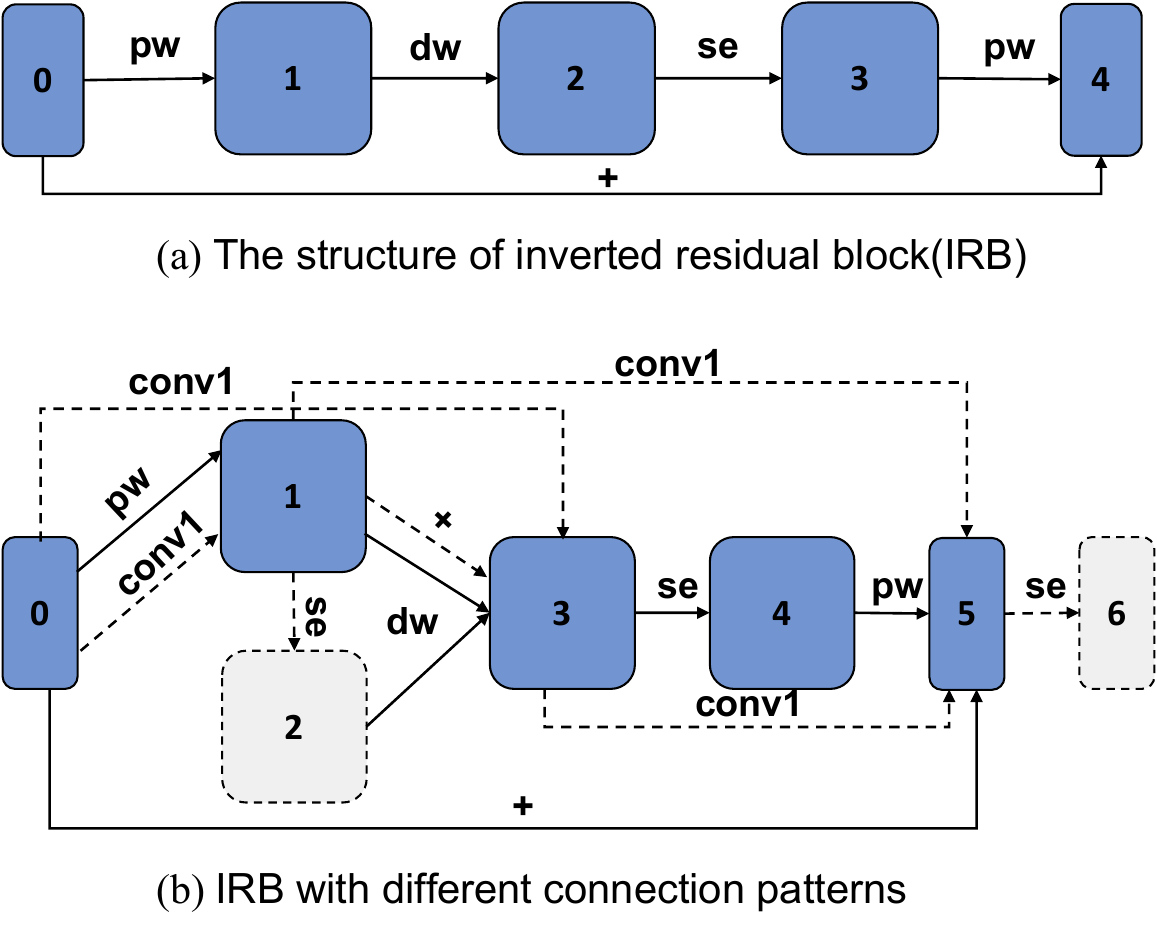}
\caption{(a): The architecture of inverted residual block(IRB), the width of the node indicates the channel numbers. (b): Our revised search space for the inverted residual block. The solid line represents the operations in original inverted residual block and the dash line represents the potential connections that can be searched in our new inverted residual block. The gray node denotes the potential node if corresponding SE operation is used. (PW: the point-wise convolution; DW: Depth-wise convolution; SE: squeeze and excite module.) }
\label{fig:imagenet_space}
\end{figure}

\begin{table}[H]
\centering 
\caption{FLOPs analysis for the IRB shown in Figure~\ref{fig:imagenet_space}(b).}
\resizebox{0.7\textwidth}{!}{
\begin{tabular}{l|r}
\toprule
\textbf{Operations between Node i and j($N_{ij}$)}  & \textbf{\#FLOPs}            \\ \midrule
point-wise-conv1($N_{01}$)              & $h_{irb} * w_{irb} * c_{m} * c_{io}$   \\ \hline
depth-wise-convk($N_{13}$)              & $h_{irb} * w_{irb} * c_{m} * k^{2}$ \\ \hline
point-wise-conv1($N_{45}$)              & $h_{irb} * w_{irb} * c_{m} * c_{io}$   \\ \hline
grouped-conv1($N_{01}$)                 & $h_{irb} * w_{irb} * c_{m}$         \\ \hline
grouped-conv1($N_{03}$)                 & $h_{irb} * w_{irb} * c_{m}$         \\ \hline
grouped-conv1($N_{15}$)                 & $h_{irb} * w_{irb} * c_{m}$        \\ \hline
grouped-conv1($N_{35}$)                 & $h_{irb} * w_{irb} * c_{m}$         \\ \hline
se-module~\cite{semodule}($N_{12}$)     & $h_{irb} * w_{irb} * c_{m} * 2r$        \\ \hline
se-module~\cite{semodule}($N_{34}$)     & $h_{irb} * w_{irb} * c_{m} * 2r$        \\ \hline
se-module~\cite{semodule}($N_{56}$)     & $h_{irb} * w_{irb} * c_{io} * 2r$        \\ \bottomrule
\end{tabular}%
}
\label{table:Flops}
\end{table}

To take the connection pattern into consideration, in this work, we redesign the connection space of the original inverted residual block (IRB) and design/add more connections, as shown in Figure~\ref{fig:imagenet_space}. Rather than only linearly connecting the input and output layer with a shortcut as in the original IRB(shown in Figure~\ref{fig:imagenet_space}(a)), we consider all possible connections inside of the IRB to form our search space(shown in Figure~\ref{fig:imagenet_space}(b)). 

In the new IRB, all nodes have new connections (dashed lines). We use a 1x1 convolution operation to connect two nodes with mismatched channel numbers and enable SE module to integrate different locations in IRB. We do this because we would let attention mechanism by SE module to be impacted on all possible connections in the IRB. 

In this paper, we searched the architecture in EfficientNet search space with this redesigned connection-searchable Inverted Residual Block using our \ours.

\subsection{Connection search space in Inverted Residual Block}
\begin{figure}[H]
\centering 
\subfloat[][Normal Cell]{\includegraphics[height=0.9in]{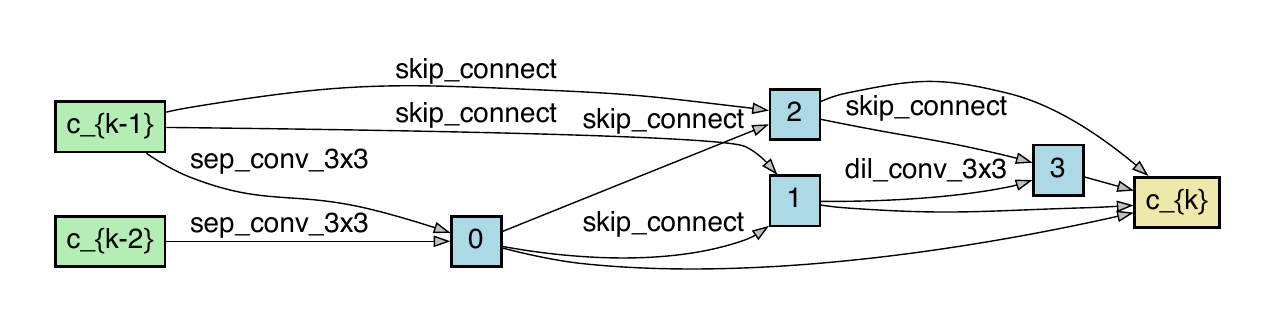}\label{normal}} \quad
\subfloat[][Reduction Cell]{\includegraphics[height=1.16in]{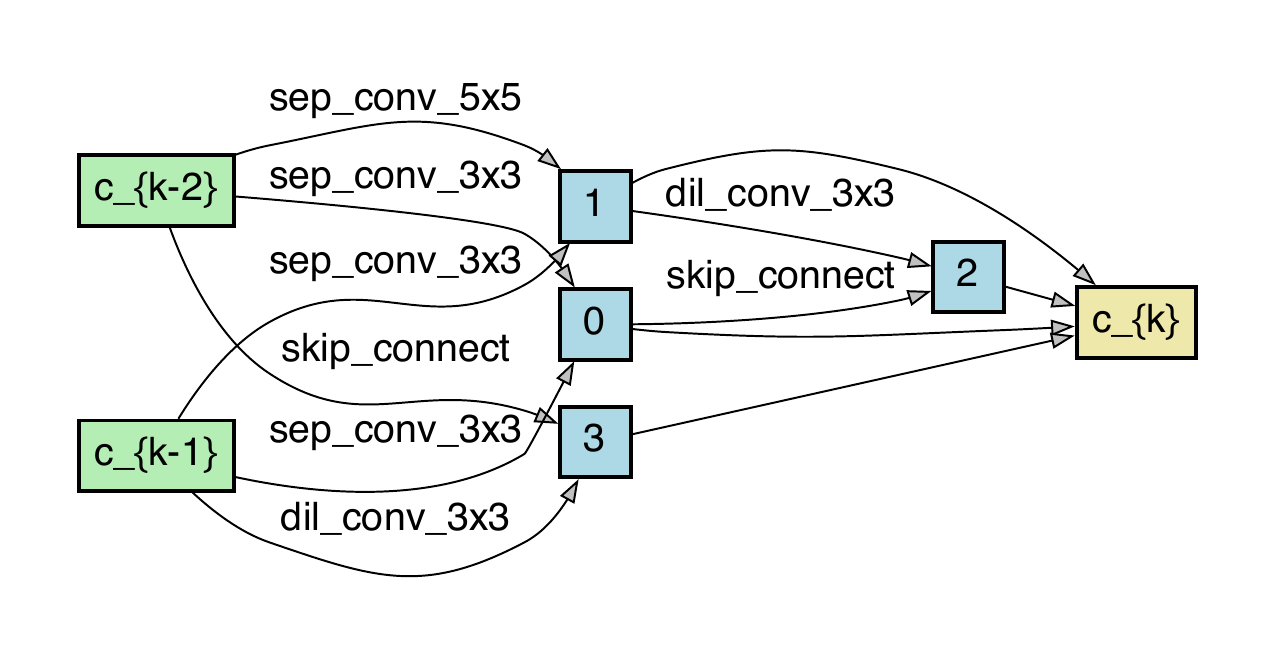}\label{reduce}}  \quad \\
\caption{Founded cell structure on Cifar10.}
\label{fig:cifar10_arch}
\end{figure}

Note that one of the key performance goals of using IRB is to achieve low computation cost. In this section, we analyze the computation cost of our modified IRB (i.e,. cost of the added connections) and compare to that of the original IRB. Our analysis shows that adding connections only increases the computation cost slightly.

\para{Cost of our new inverted residual block.}
In our search space of the connected IRB, we are only required to use 1x1 convolution to connect the mismatched channel dimensions four times as shown in Figure~\ref{fig:imagenet_space}(b). These connections correspond to edges $(N_{0}, N_{1})$, $(N_{0}, N_{3})$, $(N_{1}, N_{5})$, $(N_{3}, N_{5})$, where the subscripts correspond to the node indexes. The total cost of these 1x1 convolution operation is $4 * (h_{irb} * w_{irb} * c_{m} * c_{io})$, which is not a negligible overhead. To reduce the cost caused by 1x1 convolution, we leverage the grouped convolution~\cite{shufflenet} in place of the standard convolution. 

Specifically, we divide the output channel into number of input channel groups by first applying convolution operation in each corresponding input channel and then concatenating all the channels together. Therefore, the computational cost of these connections can be reduced to $4 * (h_{irb} * w_{irb} * c_{m})$ FLOPs if $c_{m}$ is multiple of $c_{io}$. The new squeeze and excite modules placed after first and last PW layer take $2 * h_{irb} * w_{irb} * c_{m} * r$ and $2 * h_{irb} * w_{irb} * c_{io} * r$ respectively. Note that in IRB, $c_{io}$ can be multiple times smaller than $c_{m}$. Therefore, the computational overhead associated with the added connections is bounded by $h_{irb} * w_{irb} * c_{m} * (k^{2} + 2c_{io} + 6r + 4))$.
As $c_{io}$ grows faster than both $r$ and $k$, the overhead ratio $4(r + 1) / (k^{2} + 2 * (c_{io} + r ))$ between our redesigned IRB and the original IRB is small.

\section{Details of Searched Architectures }

\begin{figure}[H]
\centering 
\includegraphics[width=0.8\columnwidth]{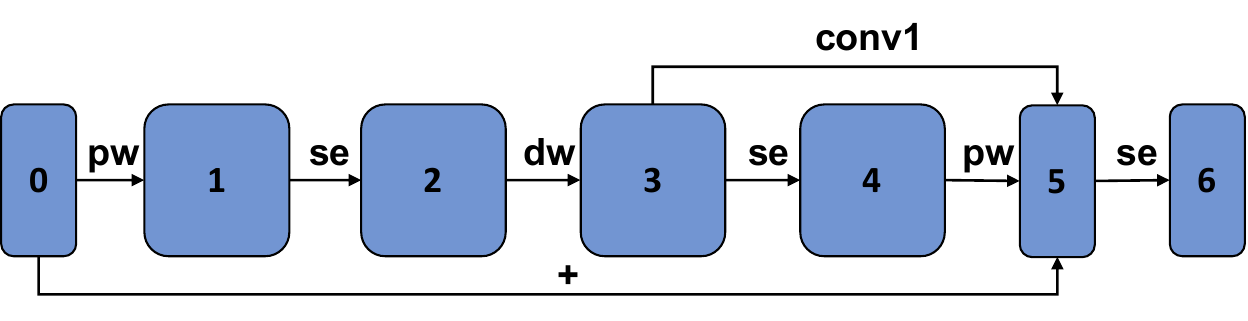}
\caption{Searched connection and SE pattern in inverted residual block}
\label{fig:searched_irb}
\end{figure}

\begin{table}[H]
\centering
 \makeatletter\def\@captype{table}\makeatother\caption{Searched ImageNet model with 248M \#FLOPs. }
 
\label{tab:imagenet:model}
\begin{tabular}{@{}cccccc@{}}
\toprule
Group                                   & Block & kernel & Stride & Channel & Expand Ratio           \\ \midrule
\multicolumn{1}{|c}{\multirow{2}{*}{0}} & Conv  & 3      & 2      & 24      & \multicolumn{1}{c|}{-}  \\
\multicolumn{1}{|c}{}                   & IRB   & 3      & 1      & 24      & \multicolumn{1}{c|}{3} \\ \midrule
\multicolumn{1}{|c}{\multirow{3}{*}{1}} & IRB   & 3      & 2      & 36      & \multicolumn{1}{c|}{3} \\
\multicolumn{1}{|c}{}                   & IRB   & 3      & 1      & 36      & \multicolumn{1}{c|}{3} \\
\multicolumn{1}{|c}{}                   & IRB   & 3      & 1      & 36      & \multicolumn{1}{c|}{3} \\ \midrule
\multicolumn{1}{|c}{\multirow{3}{*}{2}} & IRB   & 5      & 2      & 36      & \multicolumn{1}{c|}{5} \\
\multicolumn{1}{|c}{}                   & IRB   & 5      & 1      & 36      & \multicolumn{1}{c|}{5} \\
\multicolumn{1}{|c}{}                   & IRB   & 5      & 1      & 36      & \multicolumn{1}{c|}{5} \\ \midrule
\multicolumn{1}{|c}{\multirow{3}{*}{3}} & IRB   & 5      & 2      & 96      & \multicolumn{1}{c|}{5} \\
\multicolumn{1}{|c}{}                   & IRB   & 5      & 1      & 96      & \multicolumn{1}{c|}{3} \\
\multicolumn{1}{|c}{}                   & IRB   & 5      & 1      & 96      & \multicolumn{1}{c|}{3} \\ \midrule
\multicolumn{1}{|c}{\multirow{4}{*}{4}} & IRB   & 5      & 2      & 160     & \multicolumn{1}{c|}{3} \\
\multicolumn{1}{|c}{}                   & IRB   & 5      & 1      & 160     & \multicolumn{1}{c|}{6} \\
\multicolumn{1}{|c}{}                   & IRB   & 3      & 1      & 160     & \multicolumn{1}{c|}{6} \\
\multicolumn{1}{|c}{}                   & IRB   & 3      & 1      & 160     & \multicolumn{1}{c|}{7} \\ \midrule
\multicolumn{1}{|c}{\multirow{3}{*}{5}} & Conv  & 3      & 1      & 960     & \multicolumn{1}{c|}{-}  \\
\multicolumn{1}{|c}{}                   & Conv  & 1      & 1      & 1280    & \multicolumn{1}{c|}{-}  \\
\multicolumn{1}{|c}{}                   & FC    & 1      & 1      & 1000    & \multicolumn{1}{c|}{-}  \\ \bottomrule
\end{tabular}
\end{table}

The best normal and reduction cell found by \ours is visualized in Figure~\ref{fig:cifar10_arch}. Refer to ~\cite{DARTS, nasnet} for details about how to build a neural network with the searched cell. Table~\ref{tab:imagenet:model} demonstrates the founded architecture by \ours on the ImageNet dataset. Figure~\ref{fig:searched_irb} is our searched connection pattern inside of IRB.

\newpage
\bibliography{lamoo}

\end{document}

%% file: math_command.tex
\usepackage{amsmath,amsfonts,bm}

\def\eqref#1{equation~\ref{#1}}
\def\Eqref#1{Equation~\ref{#1}}

\def\1{\bm{1}}

\def\rr{{\textnormal{r}}}

\def\va{{\bm{a}}}

\def\vf{{\bm{f}}}

\def\vv{{\bm{v}}}

\def\vx{{\bm{x}}}
\def\vy{{\bm{y}}}

\DeclareMathAlphabet{\mathsfit}{\encodingdefault}{\sfdefault}{m}{sl}
\SetMathAlphabet{\mathsfit}{bold}{\encodingdefault}{\sfdefault}{bx}{n}



%% file: algorithm.tex
\algdef{SE}[SUBALG]{Indent}{EndIndent}{}{\algorithmicend\ }%
\algtext*{Indent}
\algtext*{EndIndent}

\def\root{\mathrm{root}}

\begin{algorithm*}[t]
    \small
	\caption{Pseudo-code of \oursnas for the NAS task.
    %
 }
	\label{alg:lamoo}
	\begin{algorithmic}[1]
	\State {\bfseries Inputs:} Initial $D_0$ from uniform sampling, sample budget $T$.
	\For{$t = 0, \dots, T$}
	\State Set $\mathcal{L} \leftarrow \{\Omega_\root\}$ (collections of regions to be split). 
	\While{$\mathcal{L} \neq \emptyset$}
	\State $\Omega_j \leftarrow \mathrm{pop\_first\_element}(\mathcal{L}),\ \  D_{t,j} \leftarrow D_t \cap \Omega_j, \ \ n_{t,j} \leftarrow |D_{t,j}|$. 
	\State Compute dominance number $o_{t,j}$ of $D_{t,j}$ using Eqn.~\ref{eq:dominance} and train a SVM model $h(\cdot)$.
	\State \textbf{If} $(D_{t,j}, o_{t,j})$ is splittable by SVM, \textbf{then} $\mathcal{L} \leftarrow \mathcal{L} \cup \mathrm{Partition}(\Omega_j, {h(\cdot)})$.
	\EndWhile
    \If{\emph{Path Selection}}
	\For{$k = \root$, $k$ is not leaf node}
	    \State $D_{t,k} \leftarrow D_t \cap \Omega_k, \ \ v_{t,k} \leftarrow \mathrm{HyperVolume}(D_{t,k}),\ \ n_{t,k} \leftarrow$ $|D_{t,k}|$.
	    \State $k \leftarrow \displaystyle\arg\max_{c\ \in \ \mathrm{children}(k)} \mathrm{UCB}_{t,c}$, where $\mathrm{UCB}_{t,c} := v_{t,c} + 2 C_p \sqrt{\frac{2\log(n_{t,k}}{n_{t,c}}}$
	\EndFor
	\EndIf
	\If{\emph{Leaf Selection}}
	\For{$k = \root$, $k$ is not leaf node}
	    \State $D_{t,k} \leftarrow D_t \cap \Omega_k, \ \,\ \ n_{t,k} \leftarrow$ $|D_{t,k}|$.
	\EndFor
	\EndIf
	\For{$l$ is leaf node}
	    \State $v_{t,l} \leftarrow \mathrm{HyperVolume}(D_{t,l})$
	\EndFor
	\State $k \leftarrow \displaystyle\arg\max_{l\ \in \ \mathrm{leaf\ nodes}} \mathrm{UCB}_{t,l}$, where $\mathrm{UCB}_{t,l} := v_{t,l} + 2 C_p \sqrt{\frac{2\log(n_{t,l})}{n_{t,p}}}$, where $p$ is the parent of $l$.
	
	\State $D_{t+1} \leftarrow$ $D_{t} \cup D_{\mathrm{new}}$, where $D_{\mathrm{new}}$ is drawn from $\Omega_{k}$ based on sampling algorithms such as qEHVI or CMA-ES.  
    \EndFor

  \end{algorithmic}
\end{algorithm*}